\documentclass[letterpaper]{article} % DO NOT CHANGE THIS
\usepackage[draft]{aaai25}  % DO NOT CHANGE THIS
\usepackage{times}  % DO NOT CHANGE THIS
\usepackage{helvet}  % DO NOT CHANGE THIS
\usepackage{courier}  % DO NOT CHANGE THIS
\usepackage[hyphens]{url}  % DO NOT CHANGE THIS
\usepackage{graphicx} % DO NOT CHANGE THIS
\usepackage{natbib}  % DO NOT CHANGE THIS AND DO NOT ADD ANY OPTIONS TO IT
\usepackage{caption} % DO NOT CHANGE THIS AND DO NOT ADD ANY OPTIONS TO IT
\usepackage{algorithm}
\usepackage{algorithmic}

\usepackage{newfloat}
\usepackage{listings}
\DeclareCaptionStyle{ruled}{labelfont=normalfont,labelsep=colon,strut=off} % DO NOT CHANGE THIS
\floatstyle{ruled}
\newfloat{listing}{tb}{lst}{}
\floatname{listing}{Listing}
\usepackage{xcolor}
\usepackage{microtype}
\usepackage{booktabs}
\usepackage{amsmath}
\usepackage{amssymb}
\usepackage{mathtools}
\usepackage{amsthm}
\usepackage{dsfont}
\usepackage[capitalize,noabbrev]{cleveref}
\usepackage{tikz}
\usepackage{eucal}
\tikzset{>=latex}
\usetikzlibrary{fit,automata,positioning,angles,quotes}
\usepackage{xspace}
\usepackage{subfig}

\theoremstyle{plain}
\newtheorem{theorem}{Theorem}

\theoremstyle{definition}
\newtheorem{definition}[theorem]{Definition}
\newtheorem{assumption}[theorem]{\textbf{Assumption}}
\theoremstyle{remark}
\newtheorem{remark}[theorem]{\textbf{Remark}}
\theoremstyle{definition}
\newtheorem{example}[theorem]{\textbf{Example}}
\newcommand{\hquad}{\hspace{0.5em}}
\newcommand\sguard{\diamond}
\usepackage{pifont}
\newcommand{\xmark}{\text{\ding{55}}}

\title{Progressive Safeguards for Safe and Model-Agnostic Reinforcement Learning}
\author{
    Nabil Omi\textsuperscript{\rm 1}\equalcontrib, Hosein Hasanbeig\textsuperscript{\rm 2}\equalcontrib, Hiteshi Sharma\textsuperscript{\rm 2}, Sriram K. Rajamani\textsuperscript{\rm 2}, Siddhartha Sen\textsuperscript{\rm 2}
}
\affiliations{
    \textsuperscript{\rm 1}Computer Science Department, University of Washington,~
    \textsuperscript{\rm 2}Microsoft Research
}

\usepackage{bibentry}
\begin{document}

\maketitle

\begin{abstract}
In this paper we propose a formal, model-agnostic meta-learning framework for safe reinforcement learning.
Our framework is inspired by how parents safeguard their children across a progression of increasingly riskier tasks, imparting a sense of safety that is carried over from task to task. We model this as a meta-learning process where each task is synchronized with a {\em safeguard} that monitors safety and provides a reward signal to the agent. The safeguard is implemented as a finite-state machine based on a safety specification; the reward signal is formally shaped around this specification. 
The safety specification and its corresponding safeguard can be arbitrarily complex and non-Markovian, which adds flexibility to the training process and explainability to the learned policy.
The design of the safeguard is manual but it is high-level and model-agnostic, which gives rise to an end-to-end safe learning approach with wide applicability, from pixel-level game control to language model fine-tuning. Starting from a given set of safety specifications (tasks), we train a model such that it can adapt to new specifications using only a small number of training samples. This is made possible by our method for efficiently transferring safety bias between tasks, which effectively minimizes the number of safety violations.  
We evaluate our framework in a Minecraft-inspired Gridworld, a VizDoom game environment, and an LLM fine-tuning application. Agents trained with our approach achieve near-minimal safety violations, while baselines are shown to underperform.
\end{abstract}

\section{Introduction}
Safe controller synthesis
has been studied for decades, e.g., \cite{altman1999constrained,tomlin2003computational,prajna2006barrier,romdlony2016stabilization,achiam2017constrained,fulton2018safe,le2019batch,anderson2020neurosymbolic,hunt2021verifiably,carr2022safe}, but there is still a gap between the developed theory and the challenges faced by real-world applications~\cite{amodei2016concrete}. Consider the problem of controller synthesis for a mobile robot to reach a goal while avoiding obstacles. In practice, the obstacle-free state space is often non-convex, the dynamics of the system are typically nonlinear, and the constraint and cost functions may not be convex or differentiable over all states and actions. 
%Furthermore, there are often hidden or temporal variables in the state that are not directly observable, which sometimes require non-Markovian models to be captured. 
Therefore, designing accurate models for complex real systems is challenging. In practice, low-dimensional imperfect models are typically used. Hence, the guarantees of the controllers designed using these models can be lost.

Reinforcement Learning (RL) is a controller synthesis algorithm that is widely used to train an agent to learn effective strategies for interacting with an unknown environment, modelled as a black-box Markov Decision Process (MDP). The key feature of RL is its sole dependence on a set of experience samples, which are gathered by the agent interacting with the MDP. Consequently, RL can handle problems with uncertain, stochastic, or unknown dynamics, making it well suited for problems that cannot be fully modeled. This makes RL inherently different from classical dynamic programming methods~\cite{puterman} and analytical control approaches, as RL can effectively solve the decision-making problem with minimal or no prior knowledge about the MDP~\cite{sutton}. This practical feature makes RL a great candidate for controller synthesis in areas such as economics, biology, and electrical and computer engineering, where analytical solutions are hard or infeasible to find \cite{ng,chemistry,silver}.

Despite this success, most RL exploration methods are impractical in safety-critical applications due to system vulnerabilities~\cite{garcia}. To address this shortcoming, safe RL focuses on the efficient implementation of safe exploration and policy synthesis in RL~\cite{risk1,risk2,risk3,risk4}. Traditionally, learning to behave safely in the face of uncertainty and risk is achieved either by assuming access to the system's evolution dynamics, or by predicting the future and unfolding a (learned) model of the dynamics~\cite{knownD,garcia,knownD2,shield2}. However, model-based methods suffer from model bias. They inherently assume that the learned model of the dynamics sufficiently and accurately resembles the true dynamics. Namely, a model-based agent tends to agree with itself, specifically during early stages of the learning. Model bias is especially an issue when only a few examples and no informative prior knowledge about the task are available. 

\newcommand{\tablerowheight}{0}
\begin{table*}[!ht]
\centering
\caption{Comparison of selected related work}
\label{tab:related-work}
\setlength{\tabcolsep}{0.5mm}
\scriptsize
\begin{tabular}{|llccccc|}
\hline
Approach & Method Class & Safe Exploration & Unknown Dynamics & Model Agnostic & Dynamic Safety & Scalability Reported \\[\tablerowheight pt] \hline
\cite{altman1999constrained} & Constrained Optimization & $\xmark$ & $\xmark$ & $\xmark$ & $\xmark$ & $\xmark$ \\[\tablerowheight pt] \hline
\cite{tamar2012policy} & Constrained Optimization & $\xmark$ & $\xmark$ & $\xmark$ & $\xmark$ & $\xmark$ \\[\tablerowheight pt] \hline
\cite{knownD} & Constrained Optimization & $\xmark$ & $\xmark$ & $\xmark$ & $\xmark$ & $\xmark$ \\[\tablerowheight pt] \hline
\cite{miryoosefi2019reinforcement} & Constrained Optimization & $\xmark$ & (partial) & n/a & $\xmark$ & $\xmark$ \\[\tablerowheight pt] \hline
\cite{BharadhwajKRLSG21} & Constrained Optimization & $\checkmark$ & $\checkmark$ & $\xmark$ & $\xmark$ & $\checkmark$ \\[\tablerowheight pt] \hline
\cite{paternain2022safe} & Constrained Optimization & $\xmark$ & $\checkmark$ & $\checkmark$ & $\xmark$ & $\checkmark$ \\[\tablerowheight pt] \hline
\cite{li2019temporal} & Formal Verification & $\xmark$ & $\xmark$ & $\xmark$ & $\xmark$ & $\checkmark$ \\[\tablerowheight pt] \hline
\cite{jothimurugan2020composable} & Formal Verification  & $\xmark$ & n/a & n/a & $\xmark$ & $\checkmark$ \\[\tablerowheight pt] \hline
\cite{cautiousRL} & Formal Verification  & $\checkmark$ & $\checkmark$ & $\xmark$ & $\xmark$ & $\checkmark$ \\[\tablerowheight pt] \hline
\cite{knownD2} & Gaussian Process & $\checkmark$ & (partial) & $\xmark$ & $\xmark$ & $\xmark$ \\[\tablerowheight pt] \hline
\cite{shield2} & Game Theory & $\checkmark$ & $\checkmark$ & $\xmark$ & $\xmark$ & $\xmark$ \\[\tablerowheight pt] \hline
\cite{tamar2013scaling} & Worst Case Return & $\xmark$ & $\xmark$ & $\xmark$ & $\xmark$ & $\checkmark$ \\[\tablerowheight pt] \hline
\textbf{Ours} & \textbf{Safe Meta-Learning} & $\checkmark$ & $\checkmark$ & $\checkmark$ & $\checkmark$ & $\checkmark$ \\[\tablerowheight pt] \hline
\end{tabular}
\end{table*}
\renewcommand{\arraystretch}{1}

\begin{figure}[!t]
        \centering
        \includegraphics[width=0.7\linewidth]{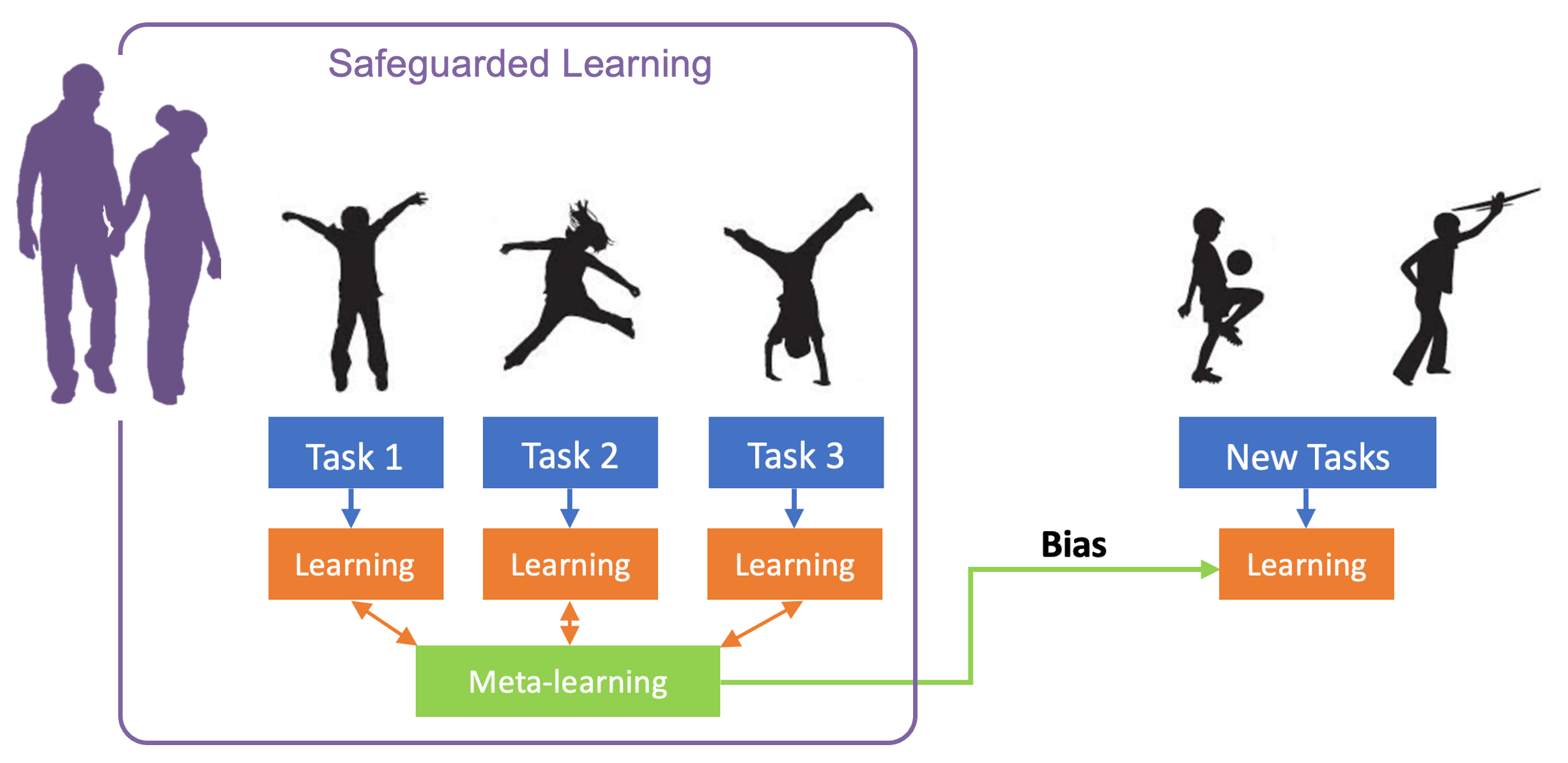}
        \caption{Transfer of safety bias in parent-child interplay.}
        \label{fig:meta_learning}
\end{figure}

To address model bias, we propose a safe meta-learning architecture that efficiently transfers the safety bias from one task to another (Figure~\ref{fig:meta_learning}) instead of employing a (potentially uninformative) model. Interestingly, there is also evidence that shows humans learn in a similar manner~\cite{wang2018prefrontal}. 
%For instance, consider the parent-child interplay: every parent raises their children in their own way, but perhaps one common aspect is that they prepare their children for the risks they might face in the real-world. This is a delicate training process in which parents expose their children to various risk-controlled activities so they become independent and adaptive to the real-world undiscovered risks. We should be able to do the same with artificial agents, training them to adapt safely and quickly from examples, and continuing to adapt as more tasks become available.  Furthermore, the definition of safety specifications in this work is high-level and model-agnostic, analogous to a parent providing safety parameters despite not knowing how to perform a task themselves.  This gives rise to an end-to-end safe learning approach with a wide range of applications, from pixel-level control to language model fine-tuning.
Parents prepare their children for real-world risks through controlled exposure, helping them become independent and adaptive. Similarly, we can train artificial agents to safely and quickly adapt to new tasks, even as more challenges arise. The safety specifications in this training are high-level and model-agnostic, much like how parents set safety parameters without knowing every task themselves. This approach enables a broad application of safe learning, from pixel-level control to language model fine-tuning.

Like most existing approaches, we require some prior high-level knowledge about unsafe elements in the environment, e.g., detectability of such elements, in order to derive the safety specifications. We introduce a stochastic safeguard that formally shapes the reward for each safety specification. The safeguard acts as a proxy between the user and the artificial agent. Specifically, a safeguard is a run-time finite-state machine that oversees the sequence of visited states and automatically checks this sequence against a safety specification. Therefore, the reward can be formally shaped around a complex and potentially non-Markovian safety specification, which is hard, if not impossible to express with standard reward shaping methods. 

Using this automatic reward shaping procedure, the agent is able to generate a policy that is safe with respect to the given specification. In the early stages of meta-learning the safeguard provides the agent with the necessary training feedback on safety as a parent does to a child. The formal reward shaping process also ensures that the learned policy in each step abides by the safety specification prescribed in that step, leaving no room for the unintended consequences of reward misspecification. Table~\ref{tab:related-work} compares the properties of our approach to a selection of existing approaches, summarizing the discussion below.

% moved this to earlier paragraph
% \noindent \textbf{Comparison to selected work.}
% Table~\ref{tab:related-work} compares the properties of our approach to a selection of existing approaches, summarizing the discussion below.

One major limitation in safe RL is the potential for unintended consequences during the policy learning process. Inherently, RL is designed to maximize a given reward function, but this can lead to the agent discovering unexpected and potentially dangerous behaviors that were not anticipated by the designer. This is particularly true in complex and high-dimensional environments where it is difficult to anticipate all possible states and behaviors. Such algorithms that learn safe policies based on the expected return often fall into the constrained MDP framework~\cite{altman1999constrained}. However, these approaches do not minimize the number of safety violations that occur \emph{during} policy learning. A primary goal of our work is to minimize safety violations while learning a safe and optimal policy.

A recent approach to minimize safety violations during learning is to use a conservative critic~\cite{BharadhwajKRLSG21} to over-approximate the probability of safety violations from states, and to avoid unsafe states. Unlike \cite{BharadhwajKRLSG21}, where the safety requirement is static throughout the learning, we provide the agent with a curriculum of requirements (Figure~\ref{fig:meta_learning}), so that the agent can update the safety requirement as needed.
An intrinsic safety-model approach is proposed in \cite{lipton2016combating}, where the safety model is a high-level function that ranks states by their safety level. This is achieved by adding penalty to nearby states with some time radius from the unsafe state. This function is called intrinsic fear and similar to our proposed approach allows the safe learning to be model-agnostic and hence, applicable to pixel-level control problems. 

Furthermore, specifying complex safety requirements in the form of a reward function can be challenging and may not always be feasible. Safety-critical systems often require the agent to consider long-term consequences of its actions. However, traditional safe RL algorithms are typically focused on short-term rewards, which can make it difficult to incorporate (potentially non-Markovian) safety requirements into the learning process. One challenge is to design reward functions that balance various aspects of the task requirements~\cite{lcrl_tool,certified_lcrl_aij,hasanbeig2023symbolic}. Prior work at the intersection of RL and formal methods leverage temporal logic to address this shortcoming by formally specifying complex tasks and automatically shaping the reward function, e.g., \cite{lcrl,li2019temporal,lcnfq,hasanbeig2020deep,jothimurugan2020composable,plmdp,cautiousRL,deepsynth,mission_dr_exploration,safeguarded_brl}. Unlike these approaches, where the goal is to perform a single and potentially complex task (modeled by a compositional specification), the goal of our work is to perform many tasks, while satisfying safety specifications.  
Recent work~\cite{ZhangKan2022} proposed meta Q-learning to learn from a diverse set of training tasks specified as linear temporal logic properties. However, maintaining safety during exploration is not core to the meta Q-learning approach. 

\section{Learning Framework}
\begin{figure}[!t]
        \centering
        \includegraphics[width=0.65\linewidth]{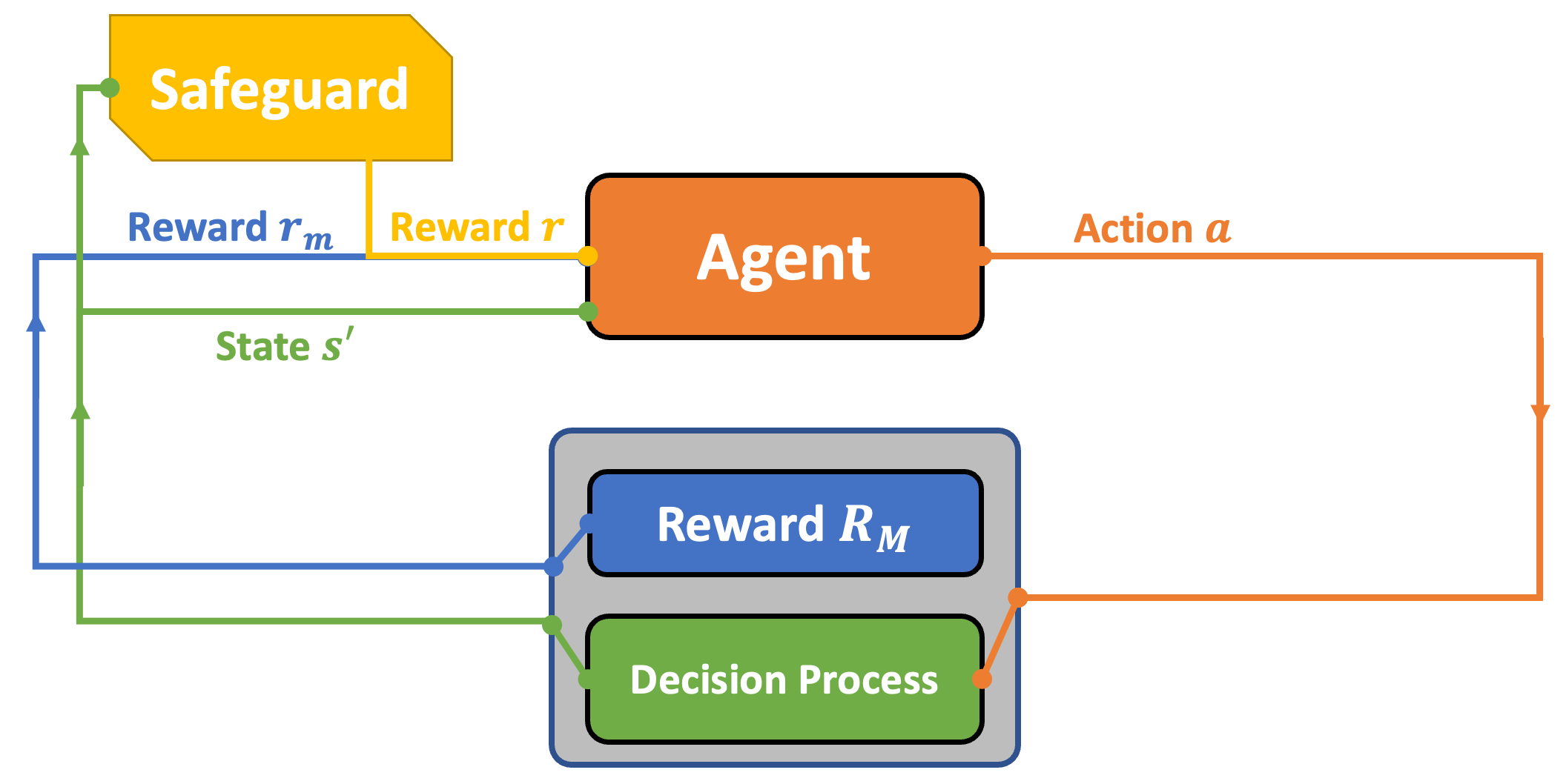}
        \caption{A simplified depiction of safeguarded learning.}
        \label{fig:architecture}
\end{figure}

In this section we elaborate on different parts of the proposed learning framework. Figure~\ref{fig:architecture} illustrates a simplified flow of the learning loop as we further discuss the underlying components in the following.
\subsection{Background on RL}
We adopt a general setting for RL, defined as follows: 
\begin{definition} [Markov Decision Process (MDP)]\label{def:mdp} 
	An MDP $\mathfrak{M}$ is a tuple $(\allowbreak\mathcal{S},\allowbreak\mathcal{A},\allowbreak \mathcal{S}_0,\allowbreak
	P)$, where $\mathcal{S}$ is the state space, $\mathcal{A}$ is a set of actions, and $\mathcal{S}_0 \subset \mathcal{S}$ is the MDP initial set of states. An initial state $s_0$ is uniformly randomly chosen from $\mathcal{S}_0$. The transition relation $P:\mathcal{S}\times\mathcal{A}\times\mathcal{B}(\mathcal{S})\rightarrow [0,1]$ is a Borel-measurable conditional kernel which assigns to any pair of state $s \in \mathcal{S}$ and action $a \in \mathcal{A}$ a probability measure $P(\cdot|s,a)$ on the Borel space $(\mathcal{S},\mathcal{B}(\mathcal{S}))$. Here, $\mathcal{B}(\mathcal{S})$ is the set of all Borel sets on $\mathcal{S}$. When the MDP has a finite state space, $P$ becomes a transition probability function, i.e., $P:\mathcal{S}\times\mathcal{A}\times\mathcal{S}\rightarrow [0,1]$.~$\hfill\square$
\end{definition}

\begin{assumption}\label{assumption_unknown_mdp}
    The general assumption in this work is that, $\mathcal{S}$, $\mathcal{A}$, and the transition relation $P$ are unknown to the agent.
\end{assumption}

A random variable $R(s,a)$ can be defined over the MDP $\mathfrak{M}$, representing the immediate reward obtained when action $a$ is taken in a given state $s$. One possible realization of this immediate reward is denoted by $r_{s,a}$.

\begin{definition}[MDP Stationary Policy]
	A policy is a rule by which the learner chooses its action at a given state in an MDP. More formally, a policy $\pi$ is a mapping from $\mathcal{S}$ to a distribution in $\mathcal{P}(\mathcal{A})$, where $\mathcal{P}(\mathcal{A})$ is the set of probability distributions on subsets of $\mathcal{A}$. A policy is stationary if $\pi(\cdot|s)\in\mathcal{P}(\mathcal{A})$ does not change over time and it is deterministic if $\pi(\cdot|s)$ is a degenerate distribution.~$\hfill\square$
\end{definition}

An MDP controlled by a policy $\pi$ induces a Markov chain $\mathfrak{M}^\pi$ with transition kernel $P^\pi(\cdot|s)=P(\cdot|s,\pi(s))$, and with reward distribution $\rho^\pi(\cdot|s)=\rho(\cdot|s,\pi(s))$ such that $R^\pi(s)\sim\rho^\pi(\cdot|s)$.

\begin{definition}
	[Expected Discounted Return] \label{def:expectedreturn} For any policy $\pi$ on an MDP $\mathfrak{M}$, the expected discounted return in state $s$ is defined as~\cite{sutton}:
	\begin{equation}
		{V}^\pi_\mathfrak{M}(s)=\mathds{E}^\pi
		\left[
		\sum\limits_{t=0}^{\infty} \gamma^t~ r_{s_t,a_t}|s_0=s
		\right],   
	\end{equation}
	where $\mathds{E}^\pi[\cdot]$ denotes the expected value by following policy $\pi$, $0\leq\gamma<1$ is the discount factor. The expected discounted return is often referred to as ``value function". Similarly, the action-value function is defined as:
	\begin{equation}\label{return}
		{Q}^\pi_\mathfrak{M}(s,a)=\mathds{E}^\pi
		\left[
		\sum\limits_{t=0}^{\infty} \gamma^t~ r_{s_t,a_t}|s_0=s,a_0=a
		\right].
	\end{equation}
	We may drop the subscript $\mathfrak{M}$ when it is clear from the context.~$\hfill\square$
\end{definition}

\begin{definition}[Optimal Policy]\label{def:optimal_pol}
	An optimal policy $\pi^*$ is defined as follows:
	$$
	\pi^*(s)=\arg\sup\limits_{\pi \in \Pi}~ {V}^\pi_\mathfrak{M}(s),
	$$
	where $\Pi$ is the set of stationary deterministic 
	policies over~$\mathcal{S}$.~$\hfill\square$
\end{definition}

\begin{theorem}[\cite{puterman}]
	In an MDP $\mathfrak{M}$ with a bounded reward function and a finite action space optimal policies are stationary and deterministic.
\end{theorem}

\subsection{Safeguard}

We introduce a formal structure that automatically shapes the reward function with respect to the specified safety specification. As mentioned before, this structure requires abstract high-level labels to be grounded over certain states. Let $\mathcal{L}$ denote the set of abstract labels, where an abstraction function $\Lambda : \mathcal{S} \times 2^{\mathcal{L}}\rightarrow [0, 1]$ specifies the agent certainty on states labels.
Specifically, $\Lambda(s,l)$ denotes the probability that the agent associates label $l\in 2^\mathcal{L}$ to state $s \in \mathcal{S}$, where $\sum_{l\in2^{\mathcal{L}}} \Lambda(s,l)=1$, $\forall s\in\mathcal{S}$. In reality an autonomous agent can determine, with some certainty, theses abstract semantics of the states. For instance, a self-driving car is able to label certain states as $\texttt{pedestrain}$ or $\texttt{moving\_car}$ with some probability, by processing all on-board sensors signals. In what follows, we discuss how these abstractions are defined over an MDP and later give rise to the safeguard definition. 

\begin{definition}[Path] \label{def:path}
In an MDP $\mathfrak{M}$, an $H$-horizon finite path $\rho$ starting at $s_0$ is a
sequence of states $\rho= s_0 \xrightarrow{a_0} s_1 \xrightarrow{a_1}\ldots{s_{H-1}} 
~$ such that every transition $s_t \xrightarrow{a_t} s_{t+1}$ is possible in
$\mathfrak{M}$, i.e.,  $s_{t+1}$ belongs to the smallest Borel set $B$ such
that $P(B|s_t,a_t)=1$.~$\hfill\square$
\end{definition}

Every path $\rho=s_0 s_1 \ldots s_{H-1}$ yields a label trace $\lambda=l_0 l_1 \ldots l_{H-1}$, where $l_t \sim \Lambda(\cdot|s_t)$ is the observed label at state $s_t$. Given the set $\mathcal{L}$, and an abstraction function $\Lambda$, a stochastic safeguard can be constructed.

\begin{definition}[Safeguard]\label{def:safe_guard} 
	A safeguard is a finite-state machine $\mathfrak{A}=(\allowbreak\mathcal{Q},\allowbreak q_0,\allowbreak\Sigma, \allowbreak\mathcal{F}, \allowbreak\delta)$ where $\mathcal{Q}$ is a finite set of states, $q_0 \in \mathcal{Q}$ is the initial state, $\Sigma=2^{\mathcal{L}}$ is the power set of abstract labels, $\mathcal{F}\subseteq\mathcal{Q}$ is the set of accepting states, and $\delta: \mathcal{Q} \times \Sigma \rightarrow \mathcal{Q}$ is a transition function.~$\hfill\square$
\end{definition}
Let $\Sigma^*$ be the set of all finite traces over $\Sigma$, and $\mathcal{Q}^*$ be the set of all finite runs, i.e., sequence of safeguard states. A finite trace $\lambda \in \Sigma^*$ is accepted by a safeguard $\mathfrak{A}$ if there exists a finite run $\theta \in \mathcal{Q}^*$ starting from $\theta[0]=q_0$ where $\theta[t+1] =\delta(\theta[t],\lambda[t]),~0\leq t < {H-1}$, and 
\begin{equation} \label{eq:acc}
\theta[H-1] \in \mathcal{F}.
\end{equation}

\begin{definition}[Rejecting Sink Component] \label{def:sinks}
	A rejecting sink component of the safeguard $\mathfrak{A}=(\mathcal{Q},q_0,\Sigma, \mathcal{F}, \delta)$ is a directed graph induced by a set of states $ Q \subset\mathcal{Q}$ such that (1) the graph is strongly connected; (2) it does not include the accepting set $\mathcal{F}$; and (3) there exist no other strongly connected set $ Q' \subset \mathcal{Q},~Q'\neq Q, $ such that $ Q \subset Q' $. We denote the union of all rejecting sink components of $\mathfrak{A}$ as $\mathcal{N}$.~$\hfill\square$
\end{definition}

Concretely, any trace $\lambda$ with associated run $\theta$ is \emph{unsafe} if at any $0 \leq t \leq H-1$
\begin{equation} \label{eq:unsafe}
\theta[t] \in \mathcal{N}.
\end{equation}

This is due to the fact that $\mathcal{N}$ is a strongly connected component in $\mathfrak{A}$, and once a run $\theta$ enters $\mathcal{N}$ there is no chance that the accepting condition \eqref{eq:acc} is met in the future. A run $\theta$ in $\mathfrak{A}$ is associated with a finite trace $\lambda$ and a path $\rho$ in the MDP $\mathfrak{M}$. 

%do not delete this line breaker
To monitor safety, the agent synchronizes the MDP state with of the safeguard to construct a ``fictitious'' MDP. Namely, as per Assumption~\ref{assumption_unknown_mdp}, this structure is not constructed in practice, and is only introduced in the following to elucidate the core concepts of the algorithm. 

\begin{definition} [Fictitious Safeguarded MDP]\label{def:guarded_mdp}
	Given an MDP $\mathfrak{M}=(\allowbreak\mathcal{S},\allowbreak\mathcal{A},\allowbreak \mathcal{S}_0,\allowbreak
	P)$ and a safeguard $\mathfrak{A}=(\allowbreak\mathcal{Q},\allowbreak q_0,\allowbreak\Sigma, \allowbreak\mathcal{F}, \allowbreak\delta)$ where $\Sigma=2^{\mathcal{L}}$, the guarded MDP (that is imagined by the agent) is defined as $$\mathfrak{M}^{\sguard}=(\mathcal{S}^{\sguard},\allowbreak \mathcal{A},\allowbreak \mathcal{S}_0^{\sguard},P^{\sguard}),$$ where $\mathcal{S}^{\sguard} = \mathcal{S}\times\mathcal{Q}$, $\mathcal{S}_0^{\sguard}=\mathcal{S}_0\times\{q_0\}$, and $P^{\sguard}:\mathcal{S}^{\sguard}\times\mathcal{A}\times\mathcal{B}(\mathcal{S}^{\sguard})\rightarrow [0,1]$ is a transition kernel such that given the current state $s^{\sguard}=(s,q)$ and action $a$, the new state is ${s^{\sguard}}'=(s',q')$, where $s'\sim P(\cdot|s,a)$ and $q'=\delta(q,l')$, where $l'\sim \Lambda(\cdot|s')$. The set $\mathcal{N}^\sguard=\mathcal{S}\times\mathcal{N}$ is the set of all unsafe states in MDP $\mathfrak{M}^{\sguard}$. When the state space is finite, then $P^{\sguard}:\mathcal{S}^{\sguard} \times \mathcal{A} \times \mathcal{S}^{\sguard} \rightarrow [0,1]$ is a transition probability function, such that : $$P^{\sguard}((s,q),a,(s',q'))=P(s,a,s')\Lambda(s',l'),$$ where $q' = \delta(q, l')$.~$\hfill\square$ 
\end{definition}

\begin{figure}[!t]
        \centering
        \includegraphics[width=0.4\linewidth]{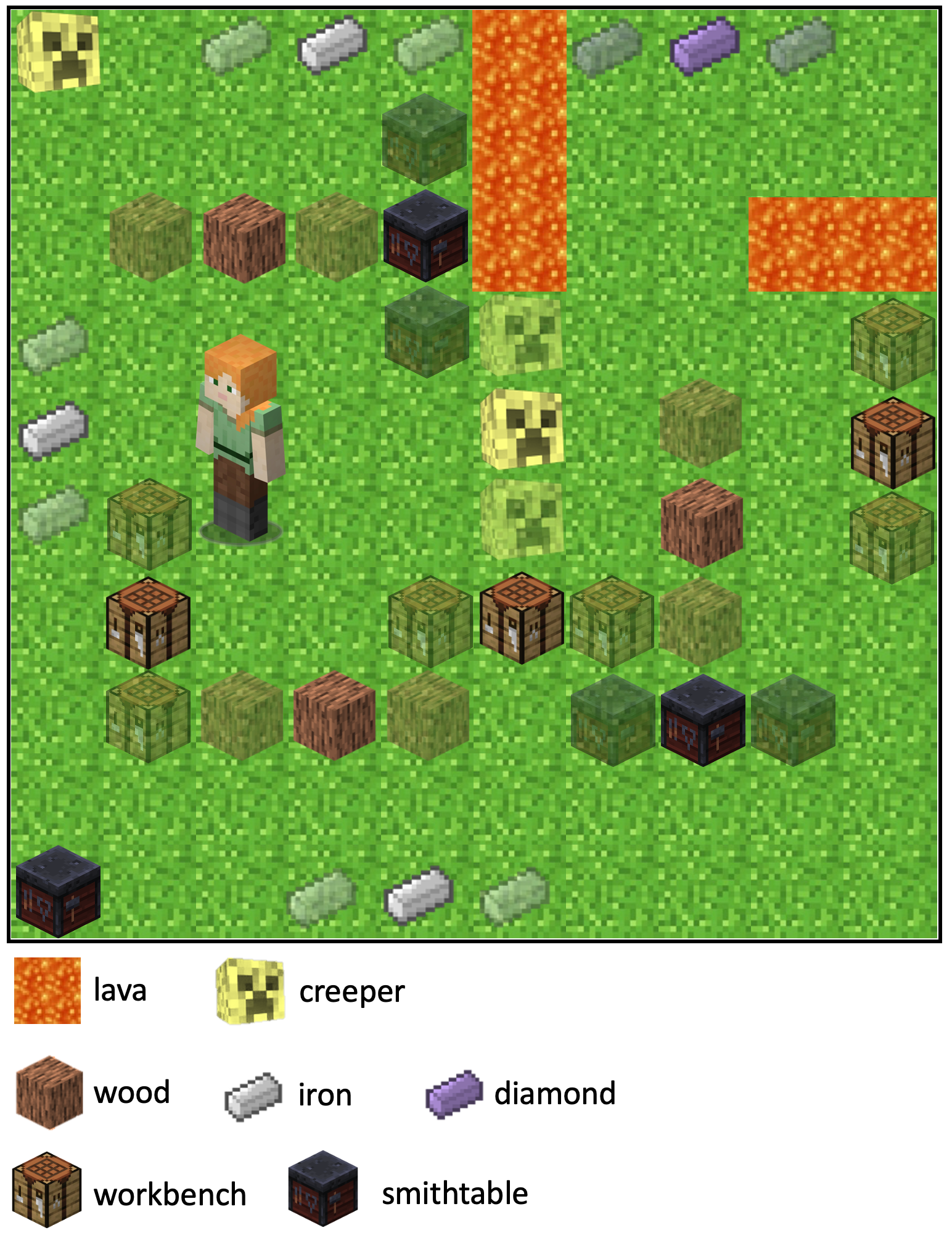}
        \caption{A stochastically-labelled Minecraft environment over which various safety specifications can be defined. The transparency level of each object corresponds to the probability of that object being observed in that location.}
        \label{fig:mine_craft}
\end{figure}

\newcommand{\scbx}{0.55}
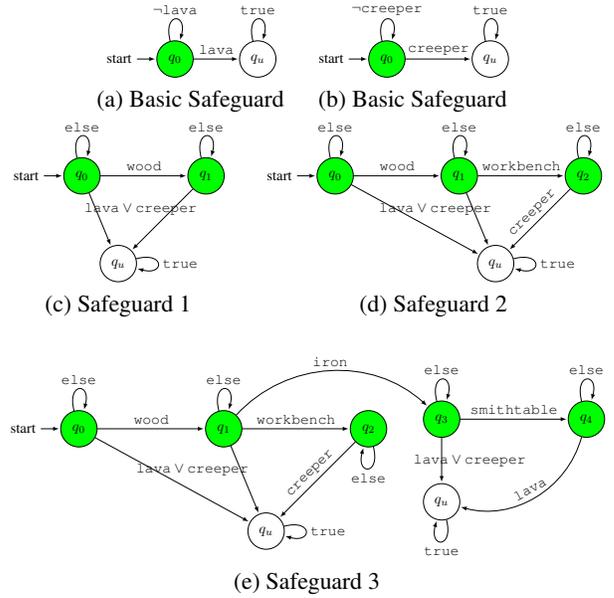
\begin{figure}[!t]\centering
\subfloat[][Basic Safeguard]{{\label{fig:building_block_1}
    \scalebox{\scbx}{
		\begin{tikzpicture}[shorten >=1pt,node distance=2cm,on grid,auto] 
		\node[state,initial,fill=green] (q_0)   {$q_0$}; 
		\node[state] (q_u) [right=of q_0]  {$q_u$}; 
		\path[->] 
        (q_0) edge [loop above] node {$\neg \texttt{lava}$} (q_0)
		(q_0) edge node {$\texttt{lava}$} (q_u)
		(q_u) edge [loop above] node {$\texttt{true}$} (q_u);
		\end{tikzpicture}
		}
		}}
\hquad
\subfloat[][Basic Safeguard]{{\label{fig:building_block_2}
    \scalebox{\scbx}{
		\begin{tikzpicture}[shorten >=1pt,node distance=2.5cm,on grid,auto] 
		\node[state,initial,fill=green] (q_0)   {$q_0$}; 
		\node[state] (q_u) [right=of q_0]  {$q_u$}; 
		\path[->] 
        (q_0) edge [loop above] node {$\neg \texttt{creeper}$} (q_0)
		(q_0) edge node {$\texttt{creeper}$} (q_u)
		(q_u) edge [loop above] node {$\texttt{true}$} (q_u);
		\end{tikzpicture}
		}
		}}
\hquad
\subfloat[][Safeguard 1]{{\label{fig:exp1_safe_guard}
    \scalebox{\scbx}{
		\begin{tikzpicture}[shorten >=1pt,node distance=3cm,on grid,auto] 
		\node[state,initial, fill=green] (q_0)   {$q_0$}; 
		\node[state,fill=green] (q_1) [right=of q_0]  {$q_1$}; 
        \node[state] (q_u) [below left=of q_1]  {$q_u$};
		\path[->] 
        (q_0) edge [loop above] node {$\texttt{else}$} (q_0)
		(q_0) edge node {$\texttt{wood}$} (q_1)
        (q_0) edge node {\hspace{-5mm}$\texttt{lava} \vee \texttt{creeper}$} (q_u)
		(q_1) edge [loop above] node {$\texttt{else}$} (q_1)
        (q_1) edge node {} (q_u)
        (q_u) edge [loop right] node {$\texttt{true}$} (q_u)
        ;
		\end{tikzpicture}
		}
		}}
	\hquad
		\subfloat[][Safeguard 2]{{\label{fig:exp2_safe_guard}
			\scalebox{\scbx}{
				\begin{tikzpicture}[shorten >=1pt,node distance=3cm,on grid,auto] 
					\node[state,initial,fill=green] (q_0)   {$q_0$}; 
					\node[state,fill=green] (q_1) [right=of q_0]  {$q_1$}; 
					\node[state,fill=green] (q_2) [right=of q_1]  {$q_2$};
					\node[state] (q_u) [below left=of q_2]  {$q_u$};
					\path[->] 
					(q_0) edge [loop above] node {$\texttt{else}$} (q_0)
					(q_0) edge node {$\texttt{wood}$} (q_1)
					(q_0) edge node {$\hspace{-10mm}\texttt{lava} \vee \texttt{creeper}$} (q_u)
					(q_1) edge [loop above] node {$\texttt{else}$} (q_1)
					(q_1) edge node {$\texttt{workbench}$} (q_2)
					(q_1) edge node {} (q_u)
					(q_2) edge [loop above] node {$\texttt{else}$} (q_2)
					(q_2) edge node [sloped] {$\texttt{creeper}$} (q_u)
					(q_u) edge [loop right] node {$\texttt{true}$} (q_u)
					;
				\end{tikzpicture}
			}
	}}
	\hquad
	\subfloat[][Safeguard 3]{{\label{fig:exp3_safe_guard}
			\scalebox{\scbx}{
				\begin{tikzpicture}[shorten >=1pt,node distance=3.5cm,on grid,auto] 
					\node[state,initial,fill=green] (q_0)   {$q_0$}; 
					\node[state,fill=green] (q_1) [right=of q_0]  {$q_1$}; 
					\node[state,fill=green] (q_2) [right=of q_1]  {$q_2$};
					\node[state] (q_d_1_3) [below right = 2.5cm of q_2] {$q_u$};
					\node[state,fill=green] (q_3) [above= 2cm of q_d_1_3]  {$q_3$};
					\node[state,fill=green] (q_4) [right=of q_3]  {$q_4$};
					\node (q_d_2_4) [above left= of q_2] {};
					\node[state] (q_u) [below left=of q_2]  {$q_u$};
					%\node[state,fill=green,accepting] (q_5) [above= 2cm of q_2]  {$q_5$};
					\path[->] 
					(q_0) edge [loop above] node {$\texttt{else}$} (q_0)
					(q_0) edge node {$\texttt{wood}$} (q_1)
					(q_0) edge node {$\hspace{-10mm}\texttt{lava} \vee \texttt{creeper}$} (q_u)
					(q_1) edge [loop above] node {$\texttt{else}$} (q_1)
					(q_1) edge node {$\texttt{workbench}$} (q_2)
					(q_1) edge [bend left=45] node {$\texttt{iron}$} (q_3)
					(q_1) edge node {} (q_u)
					(q_2) edge [loop below] node {$\texttt{else}$} (q_2)
					(q_2) edge node [sloped] {$\texttt{creeper}$} (q_u)
					%(q_2) edge node [] {$\texttt{diamond}$} (q_5)
					(q_3) edge node {$\texttt{smithtable}$} (q_4)
					(q_3) edge node {$\hspace{-8mm}\texttt{lava} \vee \texttt{creeper}$} (q_d_1_3)
					(q_3) edge [loop above] node {$\texttt{else}$} (q_3)
					(q_d_1_3) edge [loop below] node {$\texttt{true}$} (q_d_1_3)
					(q_4) edge [bend left=45] node [sloped] {$\texttt{lava}$} (q_d_1_3)
					(q_4) edge [loop above] node {$\texttt{else}$} (q_4)
					(q_u) edge [loop right] node {$\texttt{true}$} (q_u)
					;
				\end{tikzpicture}
			}
	}}
	\caption{An example of progressive safeguards. The green states are accepting states, i.e., the set $\mathcal{F}$ in Definition~\ref{def:safe_guard}. 
	An edge with label $\texttt{true}$ reads any label from the power set $2^{\mathcal{L}}$, and an edge with label $\texttt{else}$ reads any label from $2^{\mathcal{L}}$ except those that are outgoing from its node. Note that by reading labels that are unsafe with respect to the specification, the safeguard moves to a rejecting sink component (Definition~\ref{def:sinks}). As per Basic Safeguards and also Safeguard 1, interaction with $\texttt{lava}$ or $\texttt{creeper}$ is unsafe. However, Safeguard 2 allows the agent to interact with $\texttt{lava}$ after it collected $\texttt{wood}$ and went to $\texttt{workbench}$ (to create a bridge for instance). Similarly, Safeguard 3 prescribes that if the agent collects $\texttt{wood}$, $\texttt{iron}$, and goes $\texttt{smithtable}$ (to create a sword for instance), then dealing with $\texttt{creeper}$ is safe.}
	\label{fig:example_guards}
    % \vspace*{-1\baselineskip}
\end{figure}

\begin{example} [Minecraft Gridworld Environment]
\label{ex:minecraft_description}
As an example, consider the stochastically-labelled rewardless MDP in Figure~\ref{fig:mine_craft}. An agent can explore this environment with a risk-controlled set of safety specifications, e.g., ``never visit $\texttt{creeper}$'', ``do not fall into $\texttt{lava}$'', or a slightly more complicated specification ``after picking $\texttt{wood}$ interacting with $\texttt{lava}$ or $\texttt{creeper}$ is still unsafe'' (Figure~\ref{fig:example_guards}). The defined safety specifications can be arbitrarily complex over the set of abstract labels $\mathcal{L}$. 
\end{example}

By synchronizing the MDP and the safeguard as per Definition~\ref{def:guarded_mdp}, we add an extra dimension to the state space of the MDP, i.e., the states of the safeguard that represent the safety requirement. The role of the added dimension is to track safety specification satisfaction on-the-fly as RL agent explores. Note that this synchronization also converts the (potentially) non-Markovian safety policy synthesis problem over the original MDP to a Markovian one over the safeguarded MDP. In the following we elaborate on the structure of the reward function that is automatically shaped by the safeguard. 

\subsection{Safeguard-Augmented Reward}

Assume that the agent is at state $ s^\sguard=(s,q) $, takes action $ a $ and observes the subsequent state $ {s^\sguard}'=(s',q') $. Since the safeguard is a deterministic machine, $q'$ can be obtained on-the-fly. Namely, there is no need to explicitly build the safeguarded MDP and to store all its states in memory. The safeguard transitions can be executed in real-time, as the agent progresses. Given a safeguard as per Definition~\ref{def:safe_guard}, the immediate reward at state ${s^\sguard}'=(s',q')$ for taking action $a$ is a scalar value, determined by the safeguard:  
\begin{equation}\label{eq:reward}
	\begin{aligned}
		R(s^\sguard,a) = \left\{
		\begin{array}{lr}
			%r_{\mathcal{F}} + \alpha r_m&  q' \in \mathcal{F}\\
			r_{\mathcal{N}} &  q' \in \mathcal{N}\\
            r_\mathfrak{M} & \textit{otherwise}
		\end{array}
		\right.,
	\end{aligned}
\end{equation} 
where 
%$r_{\mathcal{F}}$ is the reward for visiting accepting states in the safeguard, 
$r_{\mathcal{N}}$ is the penalty for violating the safety specification by visiting the states in $\mathcal{N}$, and $r_\mathfrak{M}$ is the reward (or the penalty) received from the MDP $\mathfrak{M}$.

\begin{assumption}\label{assumption_safe_policy}
    There exists a safe policy $\Bar{\pi}$ under which the agent can reach a desired state without violating the safety requirement. This policy is unknown a priori.
\end{assumption}

\begin{definition} [Safety Probability] \label{def:safety_probab} 
	Starting from any state $s$ and following a (stationary) policy $\pi$, we denote the probability of not violating the safety specification as	
	$$
	\mathit{Pr}(\{\rho(s)\}^{\pi} \models \mathfrak{A}),
	$$
	where $\{\rho(s)\}^{\pi}$ denotes the collection of all paths starting from state $ s $, generated under policy~$\pi$. Given that we are interested in worst-case scenario, let us define the maximum probability of staying safe as:
	$$
	\mathit{Pr}_{\max}(s_0 \models\mathfrak{A})=\sup\limits_{\pi \in \Pi} \mathit{Pr}(\{\rho(s_0)\}^{\pi} \models \mathfrak{A}).
	$$
    where $s_0 \sim \textit{Uniform}(\mathcal{S}_0)$. ~$\hfill\square$
\end{definition}

\begin{remark}\label{remark:unsafety}
    Note that if the agent reads a label that leads the safeguard to $\mathcal{N}$, it is not possible to satisfy the safety specification anymore. Namely, the probability of staying safe ($\mathit{Pr}(\{\rho(s)\}^{\pi} \models \mathfrak{A})$) becomes zero under any policy $\pi\in \Pi$. Therefore, identifying $\mathcal{N}$ allows the agent to predict immediate labels that lead to a violation of the safety specification.~\hfill $\square$
\end{remark}

\begin{theorem}\label{thm:policy_optimality}
    Under Assumption~\ref{assumption_safe_policy}, for any $r_\mathcal{N}<r_\mathfrak{M}$, an optimal Markov policy $\pi^*$ on $\mathfrak{M}^\sguard$, maximizing the expected discounted return ${V}^{\pi^*}_{\mathfrak{M}^\sguard}$, maximizes the probability of not violating safety. (Proof in Appendix~A1)
\end{theorem}

Note that the optimal Markov policy $\pi^*$ on $\mathfrak{M}^\sguard$ induces a finite-memory non-Markov policy on $\mathfrak{M}$.

\section{Progressive Safeguarded Learning (PSL)}

Safe RL algorithms are traditionally designed, engineered, and tested by humans. In general, the goal of (few-shot) meta-learning is to enable an RL agent to quickly learn a policy in a new environment with limited interactions~\cite{finn2017model}. A new environment might involve accounting for a new safety specification or succeeding on a previously trained goal in a new environment. For example, we would like a mobile robot to quickly learn how to navigate through a hazardous environment so that, when in a new environment, it can determine how to reliably reach its goal safely with only a few samples. Similarly, in Example~\ref{ex:minecraft_description}, we would like the agent to quickly adapt to new safety specifications, e.g., Figure~\ref{fig:exp3_safe_guard}, after being trained on the set of safeguards in Figure~\ref{fig:building_block_1}-\ref{fig:exp2_safe_guard}. 

The task construction in this work is not random, and follows a formal curriculum. Randomly constructed tasks are often too difficult or trivial to assist the agent's learning progress. We instead propose an approach to construct tasks at the frontier of the agent's progress. This is interestingly comparable to the zone of proximal development in the brain~\cite{vygotsky2011interaction}, and the notion of scaffolding~\cite{balaban1995seeing} through which we guide the agent learning via focused interactions. The proposed safeguarded MDP structure also offers the algorithm designer flexibility to define various specifications, and also contributes to the explainability of the final solution.

% This learning structure has also been observed in large language models (LLMs), where the in-context-learning phenomenon has emerged by augmenting the prompts with related and progressive questions or answers~\cite{radford2019language}. Specifically, chain-of-thought reasoning has been found to improve the LLM answering accuracy in scenarios where several reasoning steps are required to arrive at a final answer~\cite{brown2020language}. These approaches, however, are limited by the bounded length of the input string of the LLM, and hence are restricted in many cases. More recently, novel techniques have been investigated for connecting multiple calls to a language model by processing model outputs then passing them back as subsequent inputs to the model. An example is least-to-most (LtM) prompting, where a complex reasoning question is answered first by prompting the model to produce simpler sub-questions, and passing each sub-question and resulting answer back into the model to help answer subsequent sub-questions, until eventually a final answer is reached~\cite{zhou2022least}.

% This resembles the task construction curriculum we propose in this work. 
A method for constructing similar design constraints has also been introduced in Jackson systems development (JSD)~\cite{jackson1983system}, a structured software development method that focuses on the design and implementation of visual representation of the system's structure, data flow, and control mechanisms. The JSD methodology emphasizes the importance of understanding the problem domain, data structures, and processes to create efficient and safety-reliable systems. In the running example, we started by identifying the basic safety-relevant entities in $\mathcal{L}$, i.e., $\{\texttt{lava},\texttt{creeper}\}$, and generated the basic safeguards in Figure~\ref{fig:building_block_1}-\ref{fig:building_block_2}. By identifying more safety-relevant entities, we add those entities to the previous safeguards, and gradually specify more nuanced safety requirements, e.g., Figure~\ref{fig:exp1_safe_guard}-\ref{fig:exp3_safe_guard}.

In this work, each task $\mathfrak{T}_i = \{L_{\mathfrak{T}_i}(s_0^\sguard ,\allowbreak a_0 ,\allowbreak \ldots ,\allowbreak s_{H-1}^\sguard ,\allowbreak a_{H-1} ),\allowbreak s_0, \allowbreak q_i, P^\sguard_{\mathfrak{T}_i},\allowbreak H\}$ consists of a loss function $L_{\mathfrak{T}_i}$, an initial MDP state $s_0$, a safeguard state $q_i$, a transition kernel $P^\sguard_{\mathfrak{T}_i} $, and an episode length $H$. The entire task $\mathfrak{T}_i$ can be modeled as an MDP, and thus any aspect of the MDP might change across the set of tasks $\mathfrak{T}$. In this work, each task $\mathfrak{T}_i$ corresponds to a safeguard state $q_i$ in the safeguarded MDP $\mathfrak{M}^{\sguard}_i=(\mathcal{S}^{\sguard}_i,\allowbreak \mathcal{A}_i,\allowbreak {\mathcal{S}_0^{\sguard}}_i,P^{\sguard}_i)$. In particular, for a given MDP, we only need to change the safeguards to construct and add new tasks for training. 

At each time step $t$, a parameterized policy $\pi_{\theta_i}$ maps the current state $s_t^\sguard$ to a distribution over the action space: $\pi_{\theta_i}(\cdot|s_t^\sguard)\in\mathcal{P}(\mathcal{A})$. We then define task $\mathfrak{T}_i$'s safety loss as
\begin{equation}\label{eq:loss}
	L_{\textit{safety}~\mathfrak{T}_i}(\pi_{\theta_i}) = - \mathbb{E}^{\pi_{\theta_i}}
	\left[
	\sum_{t=0}^{H-1}r_{s_t^\sguard,a_t}
	\right].
\end{equation}

The training starts with a set of safeguards that are very simple, e.g., Figures~\ref{fig:building_block_1} and \ref{fig:building_block_2}, which discourages the agent from interacting with potentially harmful objects. As the learning progresses and the agent becomes more mature, the safeguards evolve into more complex specifications. For instance, in the Minecraft example, interacting with $\texttt{lava}$ and $\texttt{creeper}$ is considered unsafe after acquiring $\texttt{wood}$ (Figure~\ref{fig:exp1_safe_guard}), but after taking $\texttt{wood}$ to $\texttt{workbench}$ the agent can go over $\texttt{lava}$ as prescribed in Figure~\ref{fig:exp2_safe_guard}.

Once a new safeguard is specified, the learned safety bias is transferred from one safeguard to another. In case when the MDP is small and a tabular approach is applicable, we transfer the safety bias explicitly but still in a model-agnostic fashion. Specifically, assume that a novel state $s^\sguard_t=(s,q_t)$ is visited and we would like the agent to generalize values across common experiences by taking the $Q$ to be a non-parametric safeguard-dependent model. Inspired by~\cite{blundell2016model,pritzel2017neural}, we approximate $Q(s^\sguard_t,a),~a\in\mathcal{A}$ by averaging the value of ancestor states of $q_t$ to construct an episodic memory:

\begin{equation}\label{eq:q_avg}
\resizebox{0.9\linewidth}{!}{$
		Q(s^\sguard_t,a) =
			\dfrac{1}{k} \sum_{i=1}^{k} \Gamma^{i-1} Q(\langle s, q_{t-i} \rangle,a),~q_{t-i} \in \textit{Anc}(q_t, k)
$}
\end{equation} 

where $\Gamma\in(0,1]$ is a bias-transfer factor, and $\textit{Anc}(q_t, k)$ is the set of (loop-free) reachable nodes from $q_t$ by repeatedly proceeding $k$-steps from child to parent in the graph of a given safeguard. Namely, $q_{t-1}$ is the closest ancestor state to $q_t$ and $q_{t-k}$ is the $k$-th ancestor. The safety bias can be transferred similarly via the parameters sets when a parameterized function approximation is used to scale the proposed architecture to large and potentially uncountably-infinite state-action decision processes. An example of such a training algorithm is outlined in Algorithm~\ref{alg:mamlrl}.

\begin{algorithm}[t]
\caption{Progressive Safeguarded Learning (PSL)}
\label{alg:mamlrl}
\begin{algorithmic}[1]
{\footnotesize
\INPUT safeguard $\mathfrak{A}$\\*
import (or initialize) $\theta_{q_i}$ for  $\pi_{\theta_{q_i}}$
\REPEAT
  \STATE initialize state $s_0$
  \FOR{$t=1,2,\ldots,H$}
      \STATE receive $s_t$
      \STATE synchronize $s_t$ with $q_t$ as per Definition~\ref{def:guarded_mdp}
      \IF{$q_t$ is never visited}
      \STATE $\theta_{q_t} \leftarrow \theta_{q_t} + \Gamma(\theta_{\textit{Anc}(q_t, 1)}-\theta_{q_t}) $
      \ENDIF
      \STATE take action $a_t$ based on $Q(s^\sguard_t,a_t|\theta_{q_t})$ \footnotesize{(e.g., Boltzmann Dist})
      \STATE receive reward $r_{s_t^\sguard,a_t}$ and next state $s^\sguard_{t+1}$
      \STATE append $(s^\sguard_t,a_t,r_{s_t^\sguard,a_t},s^\sguard_{t+1})$ to $\mathcal{D}_{q_t}$
      \STATE sample mini-batch from $\mathcal{D}_{q_t}$
      \STATE evaluate $\nabla_{\theta_{q_t}} L_{\mathfrak{T}_i}(\pi_{\theta_{q_t}})$ using $\mathcal{D}_{q_t}$ and $L_{\mathfrak{T}_i}$ as in~\eqref{eq:loss}
      \STATE gradient descent for adapted parameters:\\ $\theta_{q_t}\leftarrow\theta_{q_t}-\beta \nabla_{\theta_{q_t}}  L_{\mathfrak{T}_{q_t}}(  \pi_{\theta_{q_t}} )$
 \ENDFOR
 % \STATE update $\theta \leftarrow \theta - \beta \nabla_\theta \sum_{\{\mathfrak{T}_i\}}  L_{\mathfrak{T}_i} ( \pi_{\theta_i'})$ using each $\mathcal{D}_i'$ and $L_{\mathfrak{T}_i}$ as in~\eqref{eq:loss}
\UNTIL end of trial
}
%\STATE while 
\end{algorithmic}
\end{algorithm}

\section{Experiments}

In this section we present case studies in three domains, each posing different challenges, showcasing the flexiblity of our  proposed approach: (1) the Minecraft-inspired gridworld from Example~\ref{ex:minecraft_description}, a finite-state MDP with unknown stochastic transition dynamics; (2) the ViZDoom gaming platform~\cite{Wydmuch2019ViZdoom}, a continuous-state partially-observable MDP where observations are in the form of raw RGB video frames; and (3) an LLM fine-tuning architecture, where RL feedback is automatically shaped using safeguards, minimizing the need for a human signal. Where applicable, we compare our results with the following baselines: a standard RL approach, an intrinsic fear approach that penalizes states near in time to an unsafe state~\cite{lipton2016combating}, and a zero-shot version of our approach where the agent is only provided with the final safeguard (i.e., no progression of safeguards or meta-learning is used).  Experiment details can be found in the supplementary materials.

\newcommand{\scx}{0.45}

\begin{figure}[!t]
	\centering
    % \hspace{-7mm}
	\subfloat[][]{\includegraphics[width=\scx\linewidth]{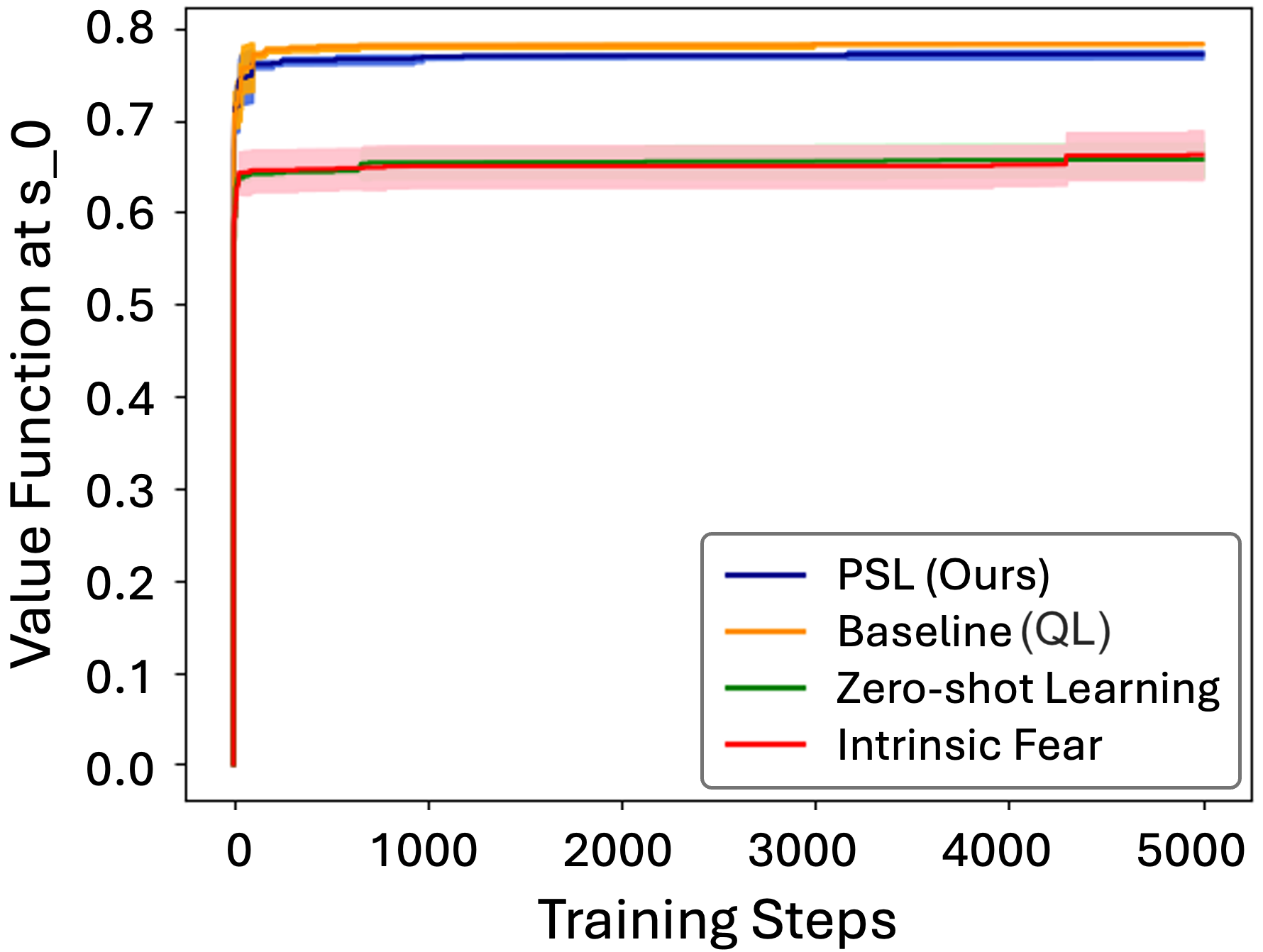}\label{fig:minecraft_convergence}}
    \quad
    \subfloat[][]{\includegraphics[width=\scx\linewidth]{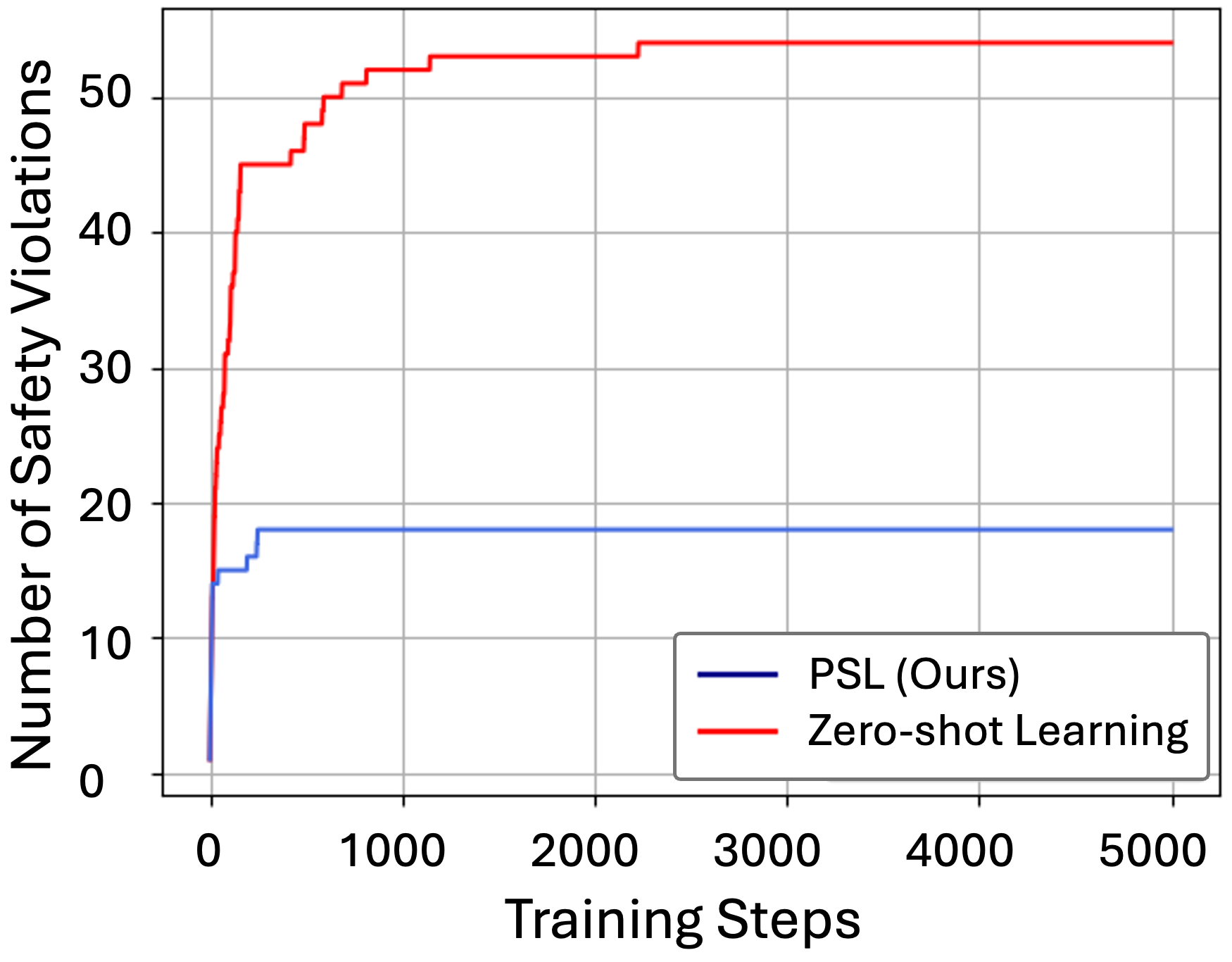}\label{fig:progressive_vs_non_progressive}}
	% \qquad
 %    \subfloat[][]{\includegraphics[width=\scx\linewidth]{fig_wood_workbench_0_unsafe_rc.png}\label{fig:exp2_policy_example}}
 %    \qquad
 %    \subfloat[][]{\includegraphics[width=\scx\linewidth]{fig_wood_workbench_iron_smithtable_0_unsafe_rc.png}\label{fig:exp3_policy_example}}
	\caption{\textbf{Minecraft experiment over 10 runs.} (a) Convergence of the expected return. (b) Cumulative number of safety violations using only Safeguard 3 (zero-shot case) compared to our approach (PSL); plots for intrinsic fear and the RL baseline are omitted for a clearer comparison, as they incur orders of magnitude more violations. 
 % (c) Policy execution after training with Safeguard 2 and (d) Safeguard 3 from Figure~\ref{fig:exp3_safe_guard}. The Minecraft map is abstracted for clarity.
 }
	\label{fig:minecraft_example_policies}
\end{figure}

% \subsection{Minecraft-inspired Gridworld}
\textbf{Minecraft-inspired Gridworld.} 
In the Minecraft example, the action space is $\mathcal{A} = \{\textit{move-north},\allowbreak\textit{move-west},\allowbreak\textit{move-south},\allowbreak\textit{move-east},\allowbreak\textit{stay}\}$. At each time step, if the agent selects action $a \in \mathcal{A}$, then there is a $95\%$ chance of moving in direction $a$, and a $5\%$ chance of moving in a random direction. If the agent ever reaches the map border and takes an action that would lead outside, then the agent will just remain in place.

Initially, we present two simple safeguards to the agent, i.e., the basic safeguards in Figure~\ref{fig:building_block_1}-\ref{fig:building_block_2}. These initial safeguards are later extended by three other safeguards outlined in Figure~\ref{fig:exp1_safe_guard}-\ref{fig:exp3_safe_guard}, forming a progression. The basic safeguards discourage the agent from interacting with any unsafe objects in the environment. Namely, the safety specifications are ``never visit $\texttt{creeper}$'', and ``do not fall into $\texttt{lava}$''. While adhering to rigid safety specifications keeps the agent safe, it prevents the agent from properly exploring the state space. Subsequent safeguards allow the agent to explore further once it learns how to safely interact with the environment.

% Adhering to these rigid safety specifications keeps the agent safe but prevents it from discovering parts of the state space and interacting with certain objects. Subsequent safeguards allow the agent to explore more of the state space, provided it has demonstrated its ability to adhere to the previous safeguards. 

As shown in Figure~\ref{fig:minecraft_convergence}, when the agent is only presented with the last and most complicated safeguard (the zero-shot case), 
% parts of the environment that include the most rewarding objects remain unexplored
the most rewarding areas of the environment are not explored, yielding a lower expected return. The intrinsic fear baseline has similarly lower performance.
Figure~\ref{fig:progressive_vs_non_progressive} shows that the average number of safety violations incurred in the zero-shot case is also higher. We omit intrinsic fear and RL baselines as they incur orders of magnitude more violations. In contrast, our approach (PSL) achieves an expected return competitive with the RL baseline, while incurring significantly fewer violations. 
% It does so by leveraging our meta-learning framework over the full progression of safeguards. 

Specifically, once the expected return (value function) at the initial state converges for a safeguard (Figure~\ref{fig:minecraft_convergence}), the agent is able to adhere to that safeguard and its safety specification. This means that for the next safeguard and its corresponding states, we can efficiently transfer the safety bias as discussed in the PSL section. As this process continues, 
% transferring the safety bias from one safeguard to the next, 
the agent is able to safely interact with objects that were previously deemed unsafe. 
An example is available in Figure~\ref{fig:exp3_policy_example_appndx} of the Appendix, where the agent safely interacts with $\texttt{lava}$ after collecting materials from the environment.
% This is evident in the policy executions depicted in  in the Appendix, where the agent safely interacts with the initially-unsafe $\texttt{lava}$ after collecting the right materials from the environment ($\texttt{wood}$ and $\texttt{workbench}$). 
% Overall, this gradual learning process seems to be more sample efficient and, more importantly, results in fewer safety violations. 

\begin{figure}[!t]
	\centering
    \subfloat[][]{\includegraphics[width=0.17\linewidth]{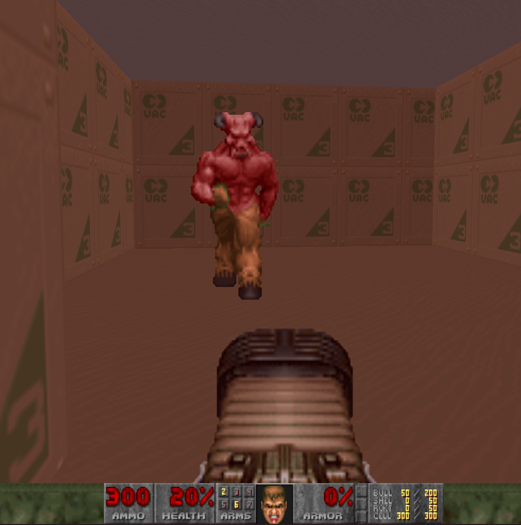}\label{fig:vizdoom_example}}
    % \quad
    \subfloat[][]{\includegraphics[width=0.7\linewidth]{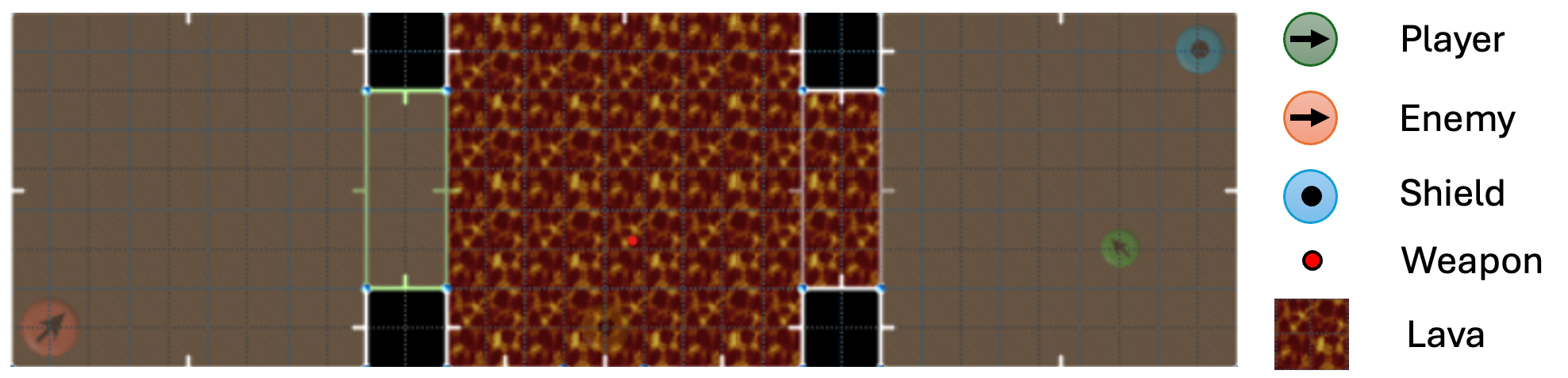}
	\label{fig:vizdoom_map}
    }
    \quad
	\subfloat[][]{\includegraphics[width=0.48\linewidth]{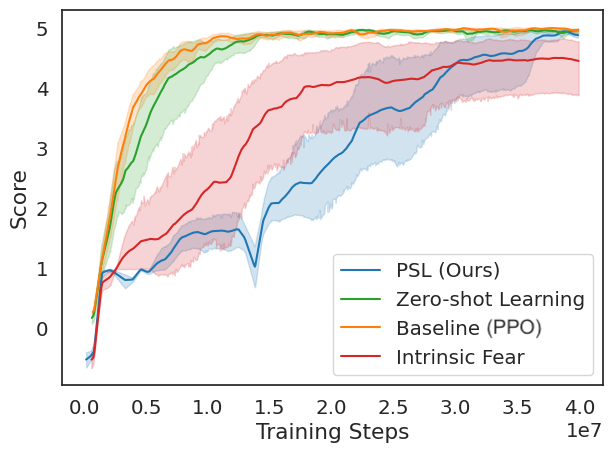}\label{fig:vizdoom_reward}}
    \quad
    \subfloat[][]{\includegraphics[width=0.47\linewidth]{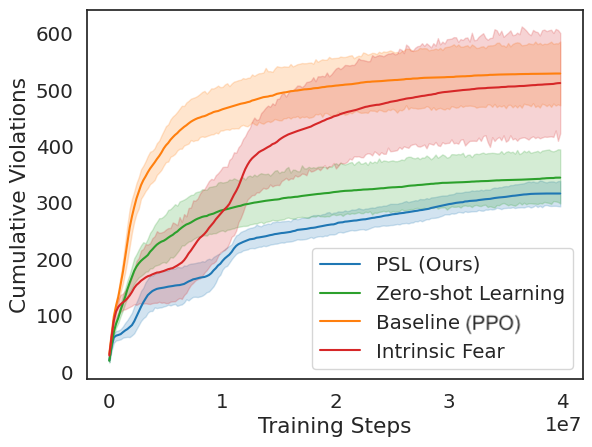}\label{fig:vizdoom_safety_violation}}
	
 \caption{\textbf{ViZDoom experiment over 10 runs.} (a) Sample frame of agent exploration. (b) Bird-eye view of the VizDoom map. (c) Comparison of the expected reward. (d) Average cumulative number of episodic safety violations.}\label{fig:vizdoom_example_policies}
\end{figure}

% \subsection{VizDoom}
\textbf{ViZDoom.} 
ViZDoom is an open-source, customizable, visually rich environment based on the first-person shooter game Doom (Figure~\ref{fig:vizdoom_example}). The environment is designed in a manner that requires the agent to interact with items to safely accomplish a task. Figure~\ref{fig:vizdoom_map} illustrates a view of the environment map, which the agent explores using RGB observations as input from an egocentric viewpoint (Figure~\ref{fig:vizdoom_example}). This makes the problem partially-observable.

%Unlike the Minecraft example, in which there is no rewarding state, the underlying game engine in ViZDoom outputs a positive reward for collecting items and eliminating enemies. Therefore, in addition to the cumulative number of safety violations, we also investigated the learning performance in terms of the expected reward.

% In this task, interacting with $\texttt{lava}$ is unsafe unless the $\texttt{shield}$ is picked up. Interacting with $\texttt{enemy}$ is unsafe unless both $\texttt{shield}$ and $\texttt{weapon}$ (in the middle of $\texttt{lava}$) are picked up. Thus, the agent must first pick up the $\texttt{shield}$, then traverse the $\texttt{lava}$ to find the $\texttt{weapon}$, which then allows it to safely interact with the $\texttt{enemy}$. The agent receives a small reward for picking up items, and for damaging the enemy. 

The safeguards we use are presented in Appendix~A2. Here, interacting with $\texttt{lava}$ is unsafe without the $\texttt{shield}$. Opening the door to $\texttt{enemy}$ is unsafe without both $\texttt{shield}$ and $\texttt{weapon}$. Thus, the agent should pick up $\texttt{shield}$ and $\texttt{weapon}$ before damaging the enemy. There is a reward for inflicting damage, and picking up items.

We compared our proposed approach (PSL applied to APPO ~\cite{pmlr_v119_petrenko20a}) to an RL baseline (APPO with the Doom engine's reward), the intrinsic fear approach~\cite{lipton2016combating}, and the zero-shot case (using only the final safeguard). Figure~\ref{fig:vizdoom_reward} and \ref{fig:vizdoom_safety_violation} shows the performance of our approach over 10 runs for expected reward and number of safety violations, respectively. Our approach (PSL) matches the expected reward level of baseline RL while incurring the fewest safety violations of all methods tested.

%It is important to note that the comparison between RL with Safeguard and vanilla RL is not one of accuracy for state-of-the-art performance, but rather a demonstration of improved learning performance.

\renewcommand{\scx}{0.5}
\begin{figure}[!t]
	\centering
    % \quad
    \subfloat[][]{\hspace{-2mm}
 \scalebox{\scx}{
 \includegraphics[width=0.9\linewidth]{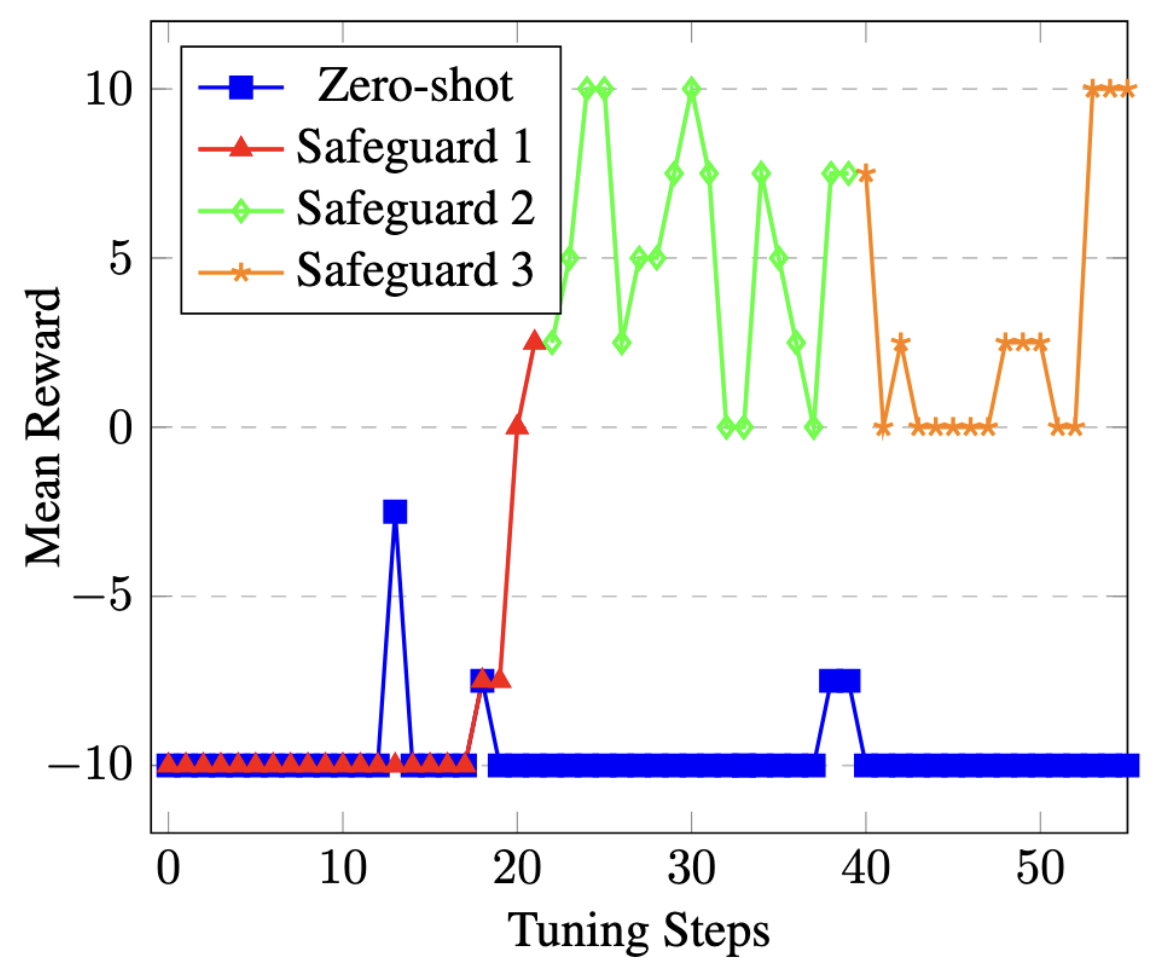}\label{fig:llm_fine_tuning_performance}}
}
    \subfloat[][]{\includegraphics[width=0.54\linewidth]{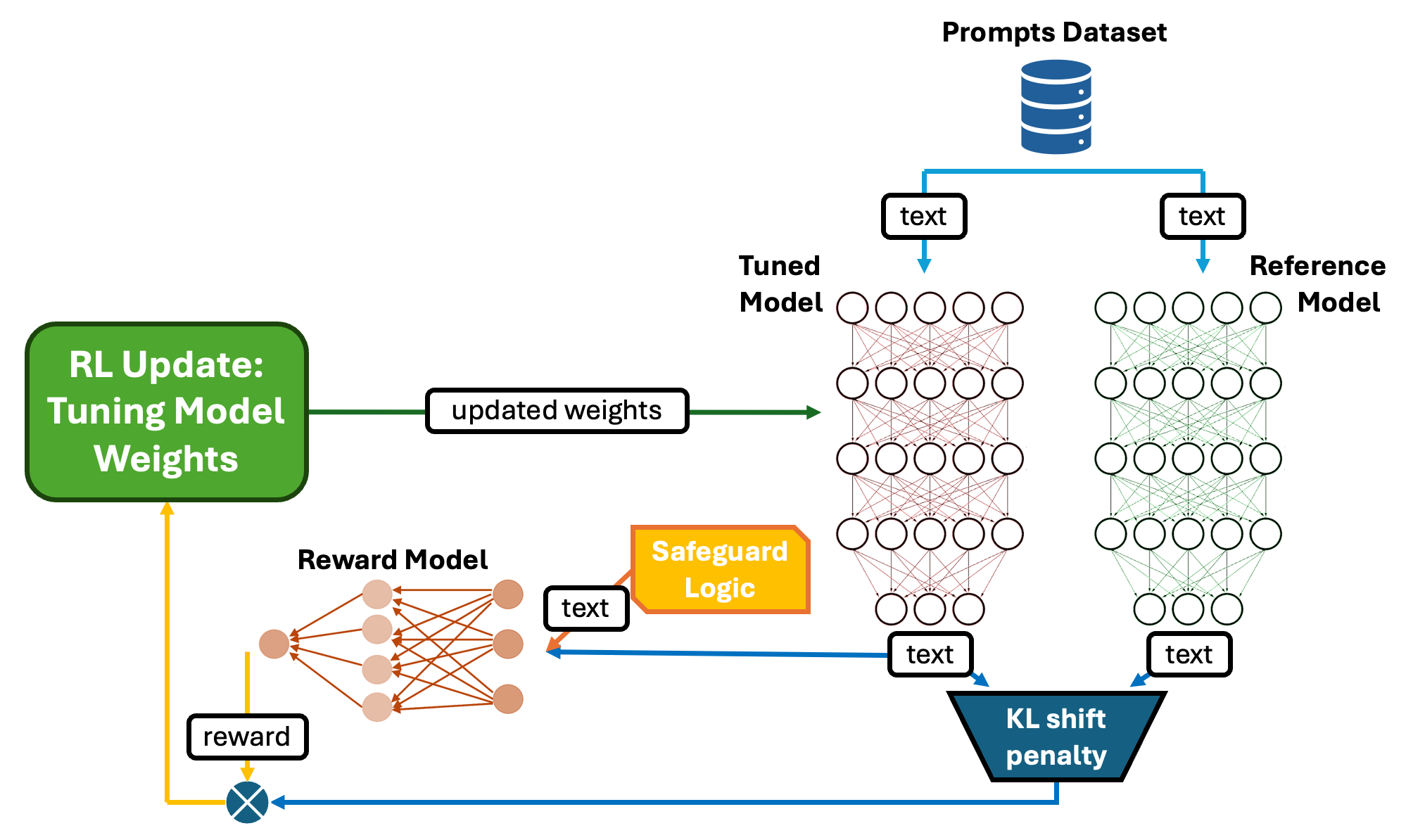}\label{fig:llm_fine_tuning}}
\caption{Fine-tuning language models using PSL. (a) Mean reward during training, (b) An overview of the RL architecture.}
\end{figure}

% \subsection{LLM Fine-tuning}
\textbf{LLM Fine-tuning.}
% Fine-tuning LLMs is another domain in which we tested the proposed approach.
% This tuning process provides a powerful approach to adapt pre-trained models for specific tasks, saving time and resources compared to training a model from scratch. However, fine-tuning requires a significant amount of
% Inspired by the success of the method in decision-making problems via RL, we leverage the expressive power of safeguards to provide such feedback in a fine-tuning RL loop
We also evaluate our approach in the domain of language model fine-tuning, where a pre-trained model is adapted to exhibit desired behavior. While fine-tuning saves time compared to training models from scratch, it requires a significant amount of labelled data and human feedback specific to the desired behavior. Inspired by fine-tuning via RLHF~\cite{ouyang2022traininglanguagemodelsfollow}, we leverage the expressive power of safeguards to provide feedback in an RL loop (Figure~\ref{fig:llm_fine_tuning}) without the need for a reward model trained on expensive human labels. The specific instructions we use are shown in Appendix~A2. We use this logic to evaluate the outputs of an LLM, and then update the parameters of the model with the resulting reward.

% While this process saves time and resources compared to training the models from scratch, fine-tuning requires a significant quantity of labeled data and human feedback specific to the desired behavior. Inspired by the success of language model fine-tuning via Reinforcement Learning from Human Feedback (RLHF) \cite{ouyang2022traininglanguagemodelsfollow}, we leverage the expressive power of safeguards to provide similar feedback in an RL loop (Figure~\ref{fig:llm_fine_tuning}). While RLHF relies on human inputs to determine a reward model, we build a reward model based on safeguard logic expressed in the form of chain-of-thought instructions;  

% In Reinforcement Learning from Human Feedback (RLHF), human input is primarily relied upon to determine the reward model. However, in this study, the reward model interprets the logic presented in the form of chain-of-thoughts instructions, and by leveraging this logic evaluates the output of the LLM. Utilizing this reward, we can then update the parameters of the LLM.

We fine-tune GPT-2 to correct CodeQL~\cite{codeql} security queries. The task is to scan the code for security vulnerabilities. We use TRL~\footnote{\texttt{github.com/huggingface/trl/tree/main}} to fine-tune the model. The reward measures how accurately the vulnerabilities are detected. Figure~\ref{fig:llm_fine_tuning_performance} shows the progress and efficiency of PSL
% 's 3-step safeguarded fine-tuning, 
compared to zero-shot fine-tuning. The safeguard logic (Figure~\ref{fig:llm_fine_tuning}) starts from a simple instruction (Safeguard 1), and as the LLM evolves we update these instructions progressively (Safeguards 2 and 3) during the fine-tuning process. 
% This allows the LLM to become proficient with the simpler instructions before moving to the harder ones. 
When using solely the final, most complicated instructions (zero-shot case), the LLM fails to achieve a meaningful reward.

\medskip
\section{Conclusion}
We introduced a formal meta-learning framework for safe reinforcement learning called Progressive Safeguarded Learning (PSL) to address the challenges of model bias, task transfer, and reward misspecification. By integrating a finite-state safeguard to protect the agent, the proposed method delivers a versatile, comprehensible, and model-agnostic solution for end-to-end safe learning. The efficacy of our approach is shown through evaluations in a Minecraft-inspired Gridworld, a ViZDoom game, and an LLM fine-tuning application. PSL achieves the same reward as existing baselines while encountering significantly fewer safety violations.
% Lastly, PSL demonstrates superior learning efficiency in certain scenarios. This is evidenced by its ability to match or exceed the performance of existing baselines while notably reducing the occurrence of safety violations.
In future work, we will investigate the potential of using PSL in conjunction with other safety mechanisms to explore the effectiveness of hybrid architectures. Additionally, we will further explore the applications of PSL in language model fine-tuning.

\bibliography{aaai25}

\begin{thebibliography}{56}
\providecommand{\natexlab}[1]{#1}

\bibitem[{Abbeel et~al.(2006)Abbeel, Coates, Quigley, and Ng}]{ng}
Abbeel, P.; Coates, A.; Quigley, M.; and Ng, A. 2006.
\newblock An application of reinforcement learning to aerobatic helicopter
  flight.
\newblock \emph{Advances in neural information processing systems}, 19.

\bibitem[{Achiam et~al.(2017)Achiam, Held, Tamar, and
  Abbeel}]{achiam2017constrained}
Achiam, J.; Held, D.; Tamar, A.; and Abbeel, P. 2017.
\newblock Constrained policy optimization.
\newblock In \emph{International conference on machine learning}, 22--31. PMLR.

\bibitem[{Altman(1999)}]{altman1999constrained}
Altman, E. 1999.
\newblock \emph{Constrained {M}arkov decision processes: stochastic modeling}.
\newblock Routledge.

\bibitem[{Amodei et~al.(2016)Amodei, Olah, Steinhardt, Christiano, Schulman,
  and Man{\'e}}]{amodei2016concrete}
Amodei, D.; Olah, C.; Steinhardt, J.; Christiano, P.; Schulman, J.; and
  Man{\'e}, D. 2016.
\newblock Concrete problems in {AI} safety.
\newblock \emph{arXiv preprint arXiv:1606.06565}.

\bibitem[{Anderson et~al.(2020)Anderson, Verma, Dillig, and
  Chaudhuri}]{anderson2020neurosymbolic}
Anderson, G.; Verma, A.; Dillig, I.; and Chaudhuri, S. 2020.
\newblock Neurosymbolic reinforcement learning with formally verified
  exploration.
\newblock \emph{Advances in neural information processing systems}, 33:
  6172--6183.

\bibitem[{Balaban(1995)}]{balaban1995seeing}
Balaban, N. 1995.
\newblock Seeing the child, knowing the person.
\newblock \emph{To become a teacher}, 52--100.

\bibitem[{Bharadhwaj et~al.(2021)Bharadhwaj, Kumar, Rhinehart, Levine, Shkurti,
  and Garg}]{BharadhwajKRLSG21}
Bharadhwaj, H.; Kumar, A.; Rhinehart, N.; Levine, S.; Shkurti, F.; and Garg, A.
  2021.
\newblock Conservative Safety Critics for Exploration.
\newblock In \emph{9th International Conference on Learning Representations,
  {ICLR} 2021, Virtual Event, Austria, May 3-7, 2021}. OpenReview.net.

\bibitem[{Blundell et~al.(2016)Blundell, Uria, Pritzel, Li, Ruderman, Leibo,
  Rae, Wierstra, and Hassabis}]{blundell2016model}
Blundell, C.; Uria, B.; Pritzel, A.; Li, Y.; Ruderman, A.; Leibo, J.~Z.; Rae,
  J.; Wierstra, D.; and Hassabis, D. 2016.
\newblock Model-free episodic control.
\newblock \emph{arXiv preprint arXiv:1606.04460}.

\bibitem[{Carr et~al.(2022)Carr, Jansen, Junges, and Topcu}]{carr2022safe}
Carr, S.; Jansen, N.; Junges, S.; and Topcu, U. 2022.
\newblock Safe Reinforcement Learning via Shielding under Partial
  Observability.
\newblock \emph{arXiv preprint arXiv:2204.00755}.

\bibitem[{codeQL(2023)}]{codeql}
codeQL. 2023.
\newblock codeQL.
\newblock https://github.com/github/codeql.

\bibitem[{Coraluppi and Marcus(1999)}]{risk1}
Coraluppi, S.~P.; and Marcus, S.~I. 1999.
\newblock Risk-sensitive and minimax control of discrete-time, finite-state
  {M}arkov decision processes.
\newblock \emph{Automatica}, 35(2): 301--309.

\bibitem[{Durrett(1999)}]{durrett1999essentials}
Durrett, R. 1999.
\newblock \emph{Essentials of stochastic processes}, volume~1.
\newblock Springer.

\bibitem[{Finn, Abbeel, and Levine(2017)}]{finn2017model}
Finn, C.; Abbeel, P.; and Levine, S. 2017.
\newblock Model-agnostic meta-learning for fast adaptation of deep networks.
\newblock In \emph{International conference on machine learning}, 1126--1135.
  PMLR.

\bibitem[{Fulton and Platzer(2018)}]{fulton2018safe}
Fulton, N.; and Platzer, A. 2018.
\newblock Safe reinforcement learning via formal methods: Toward safe control
  through proof and learning.
\newblock In \emph{Proceedings of the AAAI Conference on Artificial
  Intelligence}, volume~32.

\bibitem[{Garcia and Fern{\'a}ndez(2012)}]{risk3}
Garcia, J.; and Fern{\'a}ndez, F. 2012.
\newblock Safe exploration of state and action spaces in reinforcement
  learning.
\newblock \emph{Journal of Artificial Intelligence Research}, 45: 515--564.

\bibitem[{Garc{\i}a and Fern{\'a}ndez(2015)}]{garcia}
Garc{\i}a, J.; and Fern{\'a}ndez, F. 2015.
\newblock A comprehensive survey on safe reinforcement learning.
\newblock \emph{Journal of Machine Learning Research}, 16(1): 1437--1480.

\bibitem[{Geibel and Wysotzki(2005)}]{risk2}
Geibel, P.; and Wysotzki, F. 2005.
\newblock Risk-sensitive reinforcement learning applied to control under
  constraints.
\newblock \emph{Journal of Artificial Intelligence Research}, 24: 81--108.

\bibitem[{Hasanbeig, Abate, and Kroening(2018)}]{lcrl}
Hasanbeig, H.; Abate, A.; and Kroening, D. 2018.
\newblock Logically-Constrained Reinforcement Learning.
\newblock \emph{arXiv preprint arXiv:1801.08099}.

\bibitem[{Hasanbeig, Abate, and Kroening(2019)}]{lcnfq}
Hasanbeig, H.; Abate, A.; and Kroening, D. 2019.
\newblock Logically-Constrained Neural Fitted {Q}-Iteration.
\newblock In \emph{Proceedings of the 18th International Conference on
  Autonomous Agents and MultiAgent Systems}, 2012--2014. International
  Foundation for Autonomous Agents and Multiagent Systems.

\bibitem[{Hasanbeig, Abate, and Kroening(2020{\natexlab{a}})}]{cautiousRL}
Hasanbeig, H.; Abate, A.; and Kroening, D. 2020{\natexlab{a}}.
\newblock Cautious Reinforcement Learning with Logical Constraints.
\newblock In \emph{Proceedings of the 19th International Conference on
  Autonomous Agents and MultiAgent Systems}, 483--491. International Foundation
  for Autonomous Agents and Multiagent Systems.

\bibitem[{Hasanbeig, Abate, and
  Kroening(2020{\natexlab{b}})}]{hasanbeig2020deep}
Hasanbeig, H.; Abate, A.; and Kroening, D. 2020{\natexlab{b}}.
\newblock Deep Reinforcement Learning with Temporal Logics.
\newblock In \emph{International Conference on Formal Modeling and Analysis of
  Timed Systems}, 1--22. Springer.

\bibitem[{Hasanbeig et~al.(2019)Hasanbeig, Kantaros, Abate, Kroening, Pappas,
  and Lee}]{plmdp}
Hasanbeig, H.; Kantaros, Y.; Abate, A.; Kroening, D.; Pappas, G.~J.; and Lee,
  I. 2019.
\newblock Reinforcement Learning for Temporal Logic Control Synthesis with
  Probabilistic Satisfaction Guarantees.
\newblock In \emph{Proceedings of the 58th Conference on Decision and Control},
  5338--5343. IEEE.

\bibitem[{Hasanbeig, Kroening, and Abate(2022)}]{lcrl_tool}
Hasanbeig, H.; Kroening, D.; and Abate, A. 2022.
\newblock {LCRL}: Certified Policy Synthesis via Logically-Constrained
  Reinforcement Learning.
\newblock In \emph{International Conference on Quantitative Evaluation of
  SysTems}. Springer.

\bibitem[{Hasanbeig, Kroening, and Abate(2023)}]{certified_lcrl_aij}
Hasanbeig, H.; Kroening, D.; and Abate, A. 2023.
\newblock Certified Reinforcement Learning with Logic Guidance.
\newblock \emph{Artificial Intelligence: Special Issue on Risk-aware Autonomous
  Systems: Theory and Practice}, 103949.

\bibitem[{Hasanbeig et~al.(2021)Hasanbeig, Yogananda~Jeppu, Abate, Melham, and
  Kroening}]{deepsynth}
Hasanbeig, H.; Yogananda~Jeppu, N.; Abate, A.; Melham, T.; and Kroening, D.
  2021.
\newblock {DeepSynth}: Program Synthesis for Automatic Task Segmentation in
  Deep Reinforcement Learning.
\newblock In \emph{AAAI Conference on Artificial Intelligence}. Association for
  the Advancement of Artificial Intelligence.

\bibitem[{Hasanbeig et~al.(2024)Hasanbeig, Yogananda~Jeppu, Abate, Melham, and
  Kroening}]{hasanbeig2023symbolic}
Hasanbeig, H.; Yogananda~Jeppu, N.; Abate, A.; Melham, T.; and Kroening, D.
  2024.
\newblock Symbolic Task Inference in Deep Reinforcement Learning.
\newblock \emph{Journal of Artificial Intelligence Research (JAIR)}.

\bibitem[{Hunt et~al.(2021)Hunt, Fulton, Magliacane, Hoang, Das, and
  Solar-Lezama}]{hunt2021verifiably}
Hunt, N.; Fulton, N.; Magliacane, S.; Hoang, T.~N.; Das, S.; and Solar-Lezama,
  A. 2021.
\newblock Verifiably safe exploration for end-to-end reinforcement learning.
\newblock In \emph{Proceedings of the 24th International Conference on Hybrid
  Systems: Computation and Control}, 1--11.

\bibitem[{Jackson and Cameron(1983)}]{jackson1983system}
Jackson, M.~A.; and Cameron, J. 1983.
\newblock \emph{System development}, volume~85.
\newblock Prentice-Hall Englewood Cliffs, NJ.

\bibitem[{Jansen et~al.(2018)Jansen, K{\"o}nighofer, Junges, and
  Bloem}]{shield2}
Jansen, N.; K{\"o}nighofer, B.; Junges, S.; and Bloem, R. 2018.
\newblock Shielded decision-making in {MDPs}.
\newblock \emph{arXiv preprint arXiv:1807.06096}.

\bibitem[{Jothimurugan, Alur, and Bastani(2020)}]{jothimurugan2020composable}
Jothimurugan, K.; Alur, R.; and Bastani, O. 2020.
\newblock A Composable Specification Language for Reinforcement Learning Tasks.
\newblock arXiv:2008.09293.

\bibitem[{Le, Voloshin, and Yue(2019)}]{le2019batch}
Le, H.; Voloshin, C.; and Yue, Y. 2019.
\newblock Batch policy learning under constraints.
\newblock In \emph{International Conference on Machine Learning}, 3703--3712.
  PMLR.

\bibitem[{Li and Belta(2019)}]{li2019temporal}
Li, X.; and Belta, C. 2019.
\newblock Temporal logic guided safe reinforcement learning using control
  barrier functions.
\newblock \emph{arXiv preprint arXiv:1903.09885}.

\bibitem[{Lipton et~al.(2016)Lipton, Azizzadenesheli, Kumar, Li, Gao, and
  Deng}]{lipton2016combating}
Lipton, Z.~C.; Azizzadenesheli, K.; Kumar, A.; Li, L.; Gao, J.; and Deng, L.
  2016.
\newblock Combating reinforcement learning's sisyphean curse with intrinsic
  fear.
\newblock \emph{arXiv preprint arXiv:1611.01211}.

\bibitem[{Miryoosefi et~al.(2019)Miryoosefi, Brantley, Daume~III, Dudik, and
  Schapire}]{miryoosefi2019reinforcement}
Miryoosefi, S.; Brantley, K.; Daume~III, H.; Dudik, M.; and Schapire, R.~E.
  2019.
\newblock Reinforcement learning with convex constraints.
\newblock \emph{Advances in Neural Information Processing Systems}, 32.

\bibitem[{Mitta et~al.(2024)Mitta, Hasanbeig, Wang, Kroening, Kantaros, and
  Abate}]{safeguarded_brl}
Mitta, R.; Hasanbeig, H.; Wang, J.; Kroening, D.; Kantaros, Y.; and Abate, A.
  2024.
\newblock Safeguarded Progress in Reinforcement Learning: Safe {B}ayesian
  Exploration for Control Policy Synthesis.
\newblock In \emph{AAAI 2024}. AAAI Conference on Artificial Intelligence
  (Special Track on Safe, Robust and Responsible AI).

\bibitem[{Moldovan and Abbeel(2012)}]{knownD}
Moldovan, T.~M.; and Abbeel, P. 2012.
\newblock Safe exploration in {M}arkov decision processes.
\newblock \emph{arXiv preprint arXiv:1205.4810}.

\bibitem[{Ouyang et~al.(2022)Ouyang, Wu, Jiang, Almeida, Wainwright, Mishkin,
  Zhang, Agarwal, Slama, Ray, Schulman, Hilton, Kelton, Miller, Simens, Askell,
  Welinder, Christiano, Leike, and
  Lowe}]{ouyang2022traininglanguagemodelsfollow}
Ouyang, L.; Wu, J.; Jiang, X.; Almeida, D.; Wainwright, C.~L.; Mishkin, P.;
  Zhang, C.; Agarwal, S.; Slama, K.; Ray, A.; Schulman, J.; Hilton, J.; Kelton,
  F.; Miller, L.; Simens, M.; Askell, A.; Welinder, P.; Christiano, P.; Leike,
  J.; and Lowe, R. 2022.
\newblock Training language models to follow instructions with human feedback.
\newblock arXiv:2203.02155.

\bibitem[{Paternain et~al.(2022)Paternain, Calvo-Fullana, Chamon, and
  Ribeiro}]{paternain2022safe}
Paternain, S.; Calvo-Fullana, M.; Chamon, L.~F.; and Ribeiro, A. 2022.
\newblock Safe policies for reinforcement learning via primal-dual methods.
\newblock \emph{IEEE Transactions on Automatic Control}, 68(3): 1321--1336.

\bibitem[{Pecka and Svoboda(2014)}]{risk4}
Pecka, M.; and Svoboda, T. 2014.
\newblock Safe exploration techniques for reinforcement learning--an overview.
\newblock In \emph{International Workshop on Modelling and Simulation for
  Autonomous Systems}, 357--375. Springer.

\bibitem[{Petrenko et~al.(2020)Petrenko, Huang, Kumar, Sukhatme, and
  Koltun}]{pmlr_v119_petrenko20a}
Petrenko, A.; Huang, Z.; Kumar, T.; Sukhatme, G.; and Koltun, V. 2020.
\newblock Sample Factory: Egocentric 3{D} Control from Pixels at 100000 {FPS}
  with Asynchronous Reinforcement Learning.
\newblock In III, H.~D.; and Singh, A., eds., \emph{Proceedings of the 37th
  International Conference on Machine Learning}, volume 119 of
  \emph{Proceedings of Machine Learning Research}, 7652--7662. PMLR.

\bibitem[{Prajna(2006)}]{prajna2006barrier}
Prajna, S. 2006.
\newblock Barrier certificates for nonlinear model validation.
\newblock \emph{Automatica}, 42(1): 117--126.

\bibitem[{Pritzel et~al.(2017)Pritzel, Uria, Srinivasan, Badia, Vinyals,
  Hassabis, Wierstra, and Blundell}]{pritzel2017neural}
Pritzel, A.; Uria, B.; Srinivasan, S.; Badia, A.~P.; Vinyals, O.; Hassabis, D.;
  Wierstra, D.; and Blundell, C. 2017.
\newblock Neural episodic control.
\newblock In \emph{International conference on machine learning}, 2827--2836.
  PMLR.

\bibitem[{Puterman(2014)}]{puterman}
Puterman, M.~L. 2014.
\newblock \emph{{M}arkov decision processes: {D}iscrete stochastic dynamic
  programming}.
\newblock John Wiley \& Sons.

\bibitem[{Romdlony and Jayawardhana(2016)}]{romdlony2016stabilization}
Romdlony, M.~Z.; and Jayawardhana, B. 2016.
\newblock Stabilization with guaranteed safety using control
  {L}yapunov--barrier function.
\newblock \emph{Automatica}, 66: 39--47.

\bibitem[{Silver et~al.(2016)Silver, Huang, Maddison, Guez, Sifre, van~den
  Driessche, Schrittwieser, Antonoglou, Panneershelvam, Lanctot
  et~al.}]{silver}
Silver, D.; Huang, A.; Maddison, C.~J.; Guez, A.; Sifre, L.; van~den Driessche,
  G.; Schrittwieser, J.; Antonoglou, I.; Panneershelvam, V.; Lanctot, M.;
  et~al. 2016.
\newblock Mastering the game of {G}o with deep neural networks and tree search.
\newblock \emph{Nature}, 529(7587): 484--489.

\bibitem[{Sutton and Barto(1998)}]{sutton}
Sutton, R.~S.; and Barto, A.~G. 1998.
\newblock \emph{Reinforcement Learning: An Introduction}, volume~1.
\newblock {MIT} Press Cambridge.

\bibitem[{Tamar, Di~Castro, and Mannor(2012)}]{tamar2012policy}
Tamar, A.; Di~Castro, D.; and Mannor, S. 2012.
\newblock Policy gradients with variance related risk criteria.
\newblock In \emph{Proceedings of the twenty-ninth international conference on
  machine learning}, 387--396.

\bibitem[{Tamar, Xu, and Mannor(2013)}]{tamar2013scaling}
Tamar, A.; Xu, H.; and Mannor, S. 2013.
\newblock Scaling up robust {MDP}s by reinforcement learning.
\newblock \emph{arXiv preprint arXiv:1306.6189}.

\bibitem[{Tomlin et~al.(2003)Tomlin, Mitchell, Bayen, and
  Oishi}]{tomlin2003computational}
Tomlin, C.~J.; Mitchell, I.; Bayen, A.~M.; and Oishi, M. 2003.
\newblock Computational techniques for the verification of hybrid systems.
\newblock \emph{Proceedings of the IEEE}, 91(7): 986--1001.

\bibitem[{Turchetta, Berkenkamp, and Krause(2016)}]{knownD2}
Turchetta, M.; Berkenkamp, F.; and Krause, A. 2016.
\newblock Safe exploration in finite {M}arkov decision processes with
  {G}aussian processes.
\newblock In \emph{Advances in Neural Information Processing Systems},
  4312--4320.

\bibitem[{Vygotsky et~al.(2011)}]{vygotsky2011interaction}
Vygotsky, L.; et~al. 2011.
\newblock \emph{Interaction between learning and development}.
\newblock Link{\"o}pings universitet.

\bibitem[{Wang et~al.(2023)Wang, Hasanbeig, Tan, Sun, and
  Kantaros}]{mission_dr_exploration}
Wang, J.; Hasanbeig, H.; Tan, K.; Sun, Z.; and Kantaros, Y. 2023.
\newblock Mission-driven Exploration for Accelerated Deep Reinforcement
  Learning with Temporal Logic Task Specifications.
\newblock \emph{arXiv preprint arXiv:2311.17059}.

\bibitem[{Wang et~al.(2018)Wang, Kurth-Nelson, Kumaran, Tirumala, Soyer, Leibo,
  Hassabis, and Botvinick}]{wang2018prefrontal}
Wang, J.~X.; Kurth-Nelson, Z.; Kumaran, D.; Tirumala, D.; Soyer, H.; Leibo,
  J.~Z.; Hassabis, D.; and Botvinick, M. 2018.
\newblock Prefrontal cortex as a meta-reinforcement learning system.
\newblock \emph{Nature neuroscience}, 21(6): 860--868.

\bibitem[{Wydmuch, Kempka, and Ja\'skowski(2019)}]{Wydmuch2019ViZdoom}
Wydmuch, M.; Kempka, M.; and Ja\'skowski, W. 2019.
\newblock {ViZDoom} {C}ompetitions: {P}laying {D}oom from {P}ixels.
\newblock \emph{IEEE Transactions on Games}, 11(3): 248--259.
\newblock The 2022 IEEE Transactions on Games Outstanding Paper Award.

\bibitem[{Zhang and Kan(2022)}]{ZhangKan2022}
Zhang, H.; and Kan, Z. 2022.
\newblock Temporal Logic Guided Meta {Q}-Learning of Multiple Tasks.
\newblock \emph{IEEE Robotics and Automation Letters}, 7(3): 8194--8201.

\bibitem[{Zhou, Li, and Zare(2017)}]{chemistry}
Zhou, Z.; Li, X.; and Zare, R.~N. 2017.
\newblock Optimizing chemical reactions with deep reinforcement learning.
\newblock \emph{ACS Central Science}, 3(12): 1337--1344.

\end{thebibliography}

\appendix

\section{A1. Proof of Theorem~\ref{thm:policy_optimality}}\label{thm_proof_appndx}

\textbf{Theorem~\ref{thm:policy_optimality}.} Under Assumption~\ref{assumption_safe_policy}, for any $r_\mathcal{N}<r_\mathfrak{M}$, an optimal Markov policy $\pi^*$ on $\mathfrak{M}^\sguard$, maximizing the expected discounted return ${V}^{\pi^*}_{\mathfrak{M}^\sguard}$, maximizes the probability of not violating safety.

\begin{proof}
    Under Assumption~\ref{assumption_safe_policy}, we know that starting from a state $s^\sguard$, there exists a policy
	$\overline{\pi}$ that is safest as compared to other policies in $\Pi$. Namely, $\overline{\pi}$ satisfies $\mathfrak{A}$ with maximum (non-zero) probability (Definition~\ref{def:safety_probab}):
    \begin{equation}\label{eq:safest}
        \overline{\pi}(s^\sguard) = \arg\sup\limits_{\pi \in \Pi} \mathit{Pr}(\{\rho(s^\sguard)\}^{\pi} \models \mathfrak{A}).
    \end{equation}
    Fixing policy $\overline{\pi}$ on the MDP $\mathfrak{M}^\sguard$ induces a Markov chain ${\mathfrak{M}^\sguard}^\mathit{\overline{{\pi}}}$ by resolving the MDP action non-determinism. This induced Markov chain is a disjoint union between a set of transient states $\mathfrak{T}_{\overline{\pi}}$ and $h$ sets of irreducible recurrent classes $\mathfrak{R}^i_{\overline{\pi}},~i=1,...,h$~\cite{durrett1999essentials}, i.e., ${\mathfrak{M}^\sguard}^\mathit{\overline{{\pi}}}=\mathfrak{T}_{\overline{\pi}} \sqcup \mathfrak{R}^1_{\overline{\pi}} \sqcup ... \sqcup \mathfrak{R}^h_{\overline{\pi}}
	$. From Assumption~\ref{assumption_safe_policy} and Definition~\ref{def:guarded_mdp}, and from the fact that policy
	$\overline{\pi}$ is a safe policy, we conclude:
    \begin{equation}\label{eq:no_intersec}
        \exists \mathfrak{R}^i_{\overline{\pi}} ~s.t.~ \mathcal{N}^\sguard \cap \mathfrak{R}^i_{\overline{\pi}} = \emptyset
    \end{equation}
    This is also true for any policy $\pi$ whose traces $\{\rho(s^\sguard)\}^{\pi}$ satisfy safety $\mathfrak{A}$ with positive probability. Let us assume that an optimal policy $\pi^*$ is not the safest policy. In what follows, by contradiction, we show that any optimal policy ${\pi}^*$ which maximizes the expected discounted reward is indeed the safest policy as per \eqref{eq:safest} if $r_\mathcal{N}<r_\mathfrak{M}$. 
    
    From Definition~\ref{def:optimal_pol}, for an optimal policy $\pi^*$ we have:
    $$
    {\pi^*}(s^\sguard)=\arg\sup\limits_{\pi \in \Pi} \mathds{E}^{\pi}
		\left[
		\sum\limits_{t=0}^{\infty} \gamma^t~ r_t~|~s^\sguard_0=s^\sguard
		\right].
    $$
    The expected discounted reward can be rewritten as the expected return of the collection of satisfying and violating paths, namely:
    \begin{align}\label{eq:return_split}
    \resizebox{\columnwidth}{!}{$
	\begin{aligned}
    &{\pi^*}(s^\sguard)=\arg\sup\limits_{\pi \in \Pi}(\\
    &\mathds{E}^{\pi}
		\left[
		\sum\limits_{t=0}^{\infty} \gamma^t~ r_t~|~s^\sguard_0=s^\sguard,~\{\rho(s^\sguard)\}^{\pi} \models \mathfrak{A}
		\right] \times \mathit{Pr}(\{\rho(s^\sguard)\}^{\pi} \models \mathfrak{A}) + \\
    &\mathds{E}^{\pi}
		\left[
		\sum\limits_{t=0}^{\infty} \gamma^t~ r_t~|~s^\sguard_0=s^\sguard,~\{\rho(s^\sguard)\}^{\pi} \not\models \mathfrak{A}
		\right] \times \mathit{Pr}(\{\rho(s^\sguard)\}^{\pi} \not\models \mathfrak{A})~)
    \end{aligned}$}
    % ERROR: overfull, REMOVE
	\end{align}
    \begin{align}
	\begin{aligned}
    {\pi^*}(s^\sguard)=&\arg\sup\limits_{\pi \in \Pi}(~
		r_\mathfrak{M}/(1-\gamma) \times \mathit{Pr}(\{\rho(s^\sguard)\}^{\pi} \models \mathfrak{A}) + \\ 
		& r_\mathcal{N}/(1-\gamma) \times \mathit{Pr}(\{\rho(s^\sguard)\}^{\pi} \not\models \mathfrak{A})~)
    \end{aligned}
	\end{align}
    From $\mathit{Pr}(\{\rho(s^\sguard)\}^{\pi} \not\models \mathfrak{A}) = 1 - \mathit{Pr}(\{\rho(s^\sguard)\}^{\pi} \models \mathfrak{A})$ we can restructure the above as:
    \begin{align}
    \resizebox{\columnwidth}{!}{$
	\begin{aligned}
    {\pi^*}(s^\sguard)&=\arg\sup\limits_{\pi \in \Pi}(~
		r_\mathfrak{M}/(1-\gamma) \times \mathit{Pr}(\{\rho(s^\sguard)\}^{\pi} \models \mathfrak{A}) + \\
		&~~~ r_\mathcal{N}/(1-\gamma) \times (1 - \mathit{Pr}(\{\rho(s^\sguard)\}^{\pi} \models \mathfrak{A}))~) \\
            &=\arg\sup\limits_{\pi \in \Pi}(~
		(r_\mathfrak{M}-r_\mathcal{N})/(1-\gamma) \times \mathit{Pr}(\{\rho(s^\sguard)\}^{\pi} \models \mathfrak{A}) + \\
		&~~~ r_\mathcal{N}/(1-\gamma)~)
    \end{aligned}$}
    % ERROR, overfull, REMOVE
	\end{align}
    Since $r_\mathcal{N} < r_\mathfrak{M}$, and $\gamma<1$ (Definition~\ref{def:expectedreturn}), then
    \begin{equation}
        {\pi^*}(s^\sguard) = \arg\sup\limits_{\pi \in \Pi} \mathit{Pr}(\{\rho(s^\sguard)\}^{\pi} \models \mathfrak{A}).
    \end{equation}
    This shows that any optimal policy ${\pi}^*$ which maximizes the expected discounted reward is indeed the safest policy as per \eqref{eq:safest}.
\end{proof}

\section{A2. Details on the Experiments}

All the experiments are implemented on a machine with an AMD EPYC 64-Core CPU, Tesla T4 GPU, and 100 GB of RAM running Ubuntu 20.04.6. Instructions on running experiments can be found on the GitHub page for this paper.

\subsection{Minecraft}
\begin{figure}[!t]
	\centering
    \subfloat[][]{\includegraphics[width=0.4\linewidth]{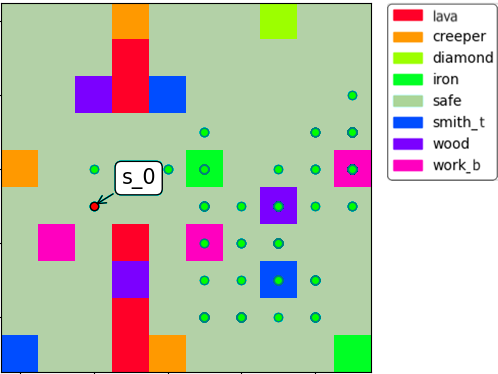}}
    \quad
    \subfloat[][]{\includegraphics[width=0.49\linewidth]{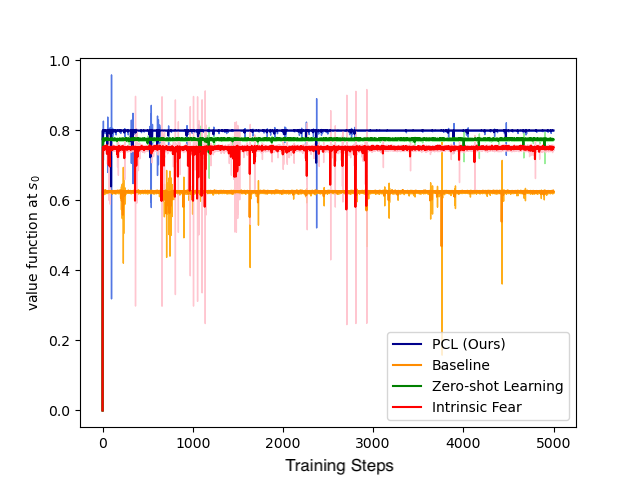}
    }
    \quad
	\subfloat[][]{\includegraphics[width=0.4\linewidth]{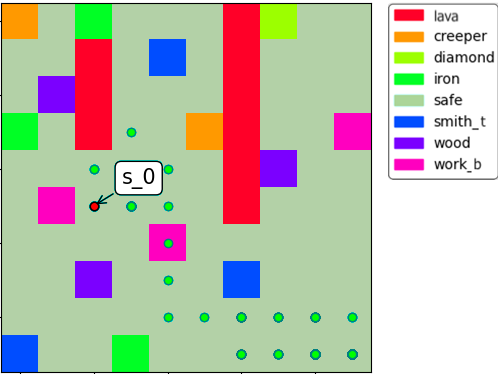}}
    \quad
    \subfloat[][]{\includegraphics[width=0.49\linewidth]{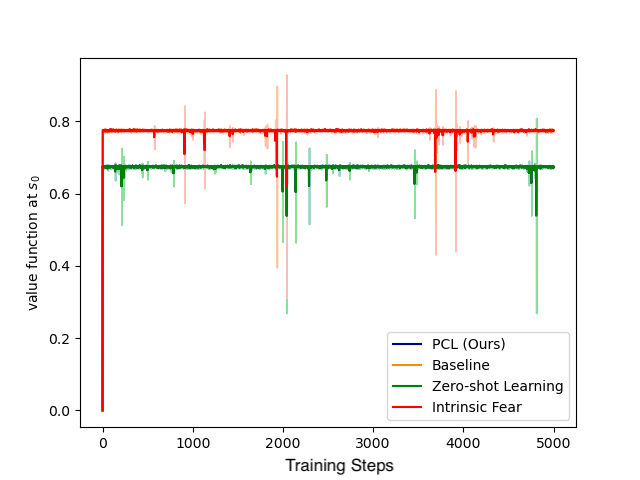}}
	
 \caption{Minecraft environment variations.}\label{fig:minecraft_layouts}
\end{figure}

As the agent becomes more mature, and by the efficient transfer of the safety bias to new safeguards, the agent can safely interact with the objects that were initially deemed as unsafe. 
This is evident in the policy executions depicted in Figure~\ref{fig:exp2_policy_example_appndx}-\ref{fig:exp3_policy_example_appndx}. 
As described before, after collecting the right ingredients in the environment, the agent learns to interact with initially-unsafe parts of the state space, e.g., $\texttt{lava}$. We also tested the performance of PSL on different layouts of Minecraft (Figure~\ref{fig:minecraft_layouts}) and PSL managed to reach the objectives more efficiently.

\renewcommand{\scx}{0.45}

\begin{figure}[!b]
	\centering
	% \subfloat[][policy execution after training with Safeguard 1 in Figure~\ref{fig:exp1_safe_guard} (map abstracted)]{\includegraphics[width=\scx\linewidth]{fig_wood_0_unsafe_rc.png}\label{fig:exp1_policy_example_appndx}}
	% \qquad
	\subfloat[][policy execution after training with Safeguard 2 in Figure~\ref{fig:exp2_safe_guard} (map abstracted)]{\includegraphics[width=\scx\linewidth]{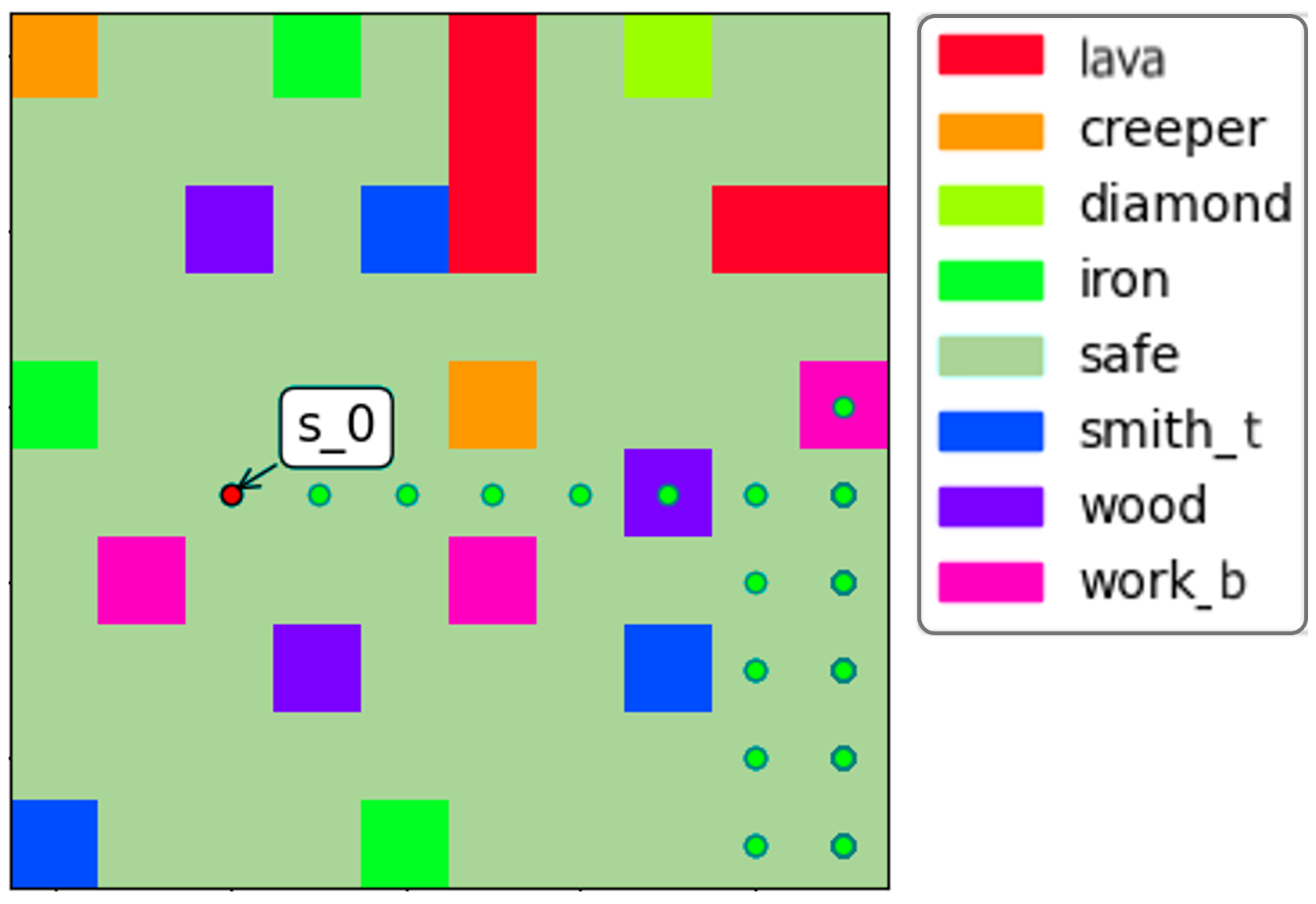}\label{fig:exp2_policy_example_appndx}}
	\qquad
	\subfloat[][policy execution after training with Safeguard 3 in Figure~\ref{fig:exp3_safe_guard} (map abstracted)]{\includegraphics[width=\scx\linewidth]{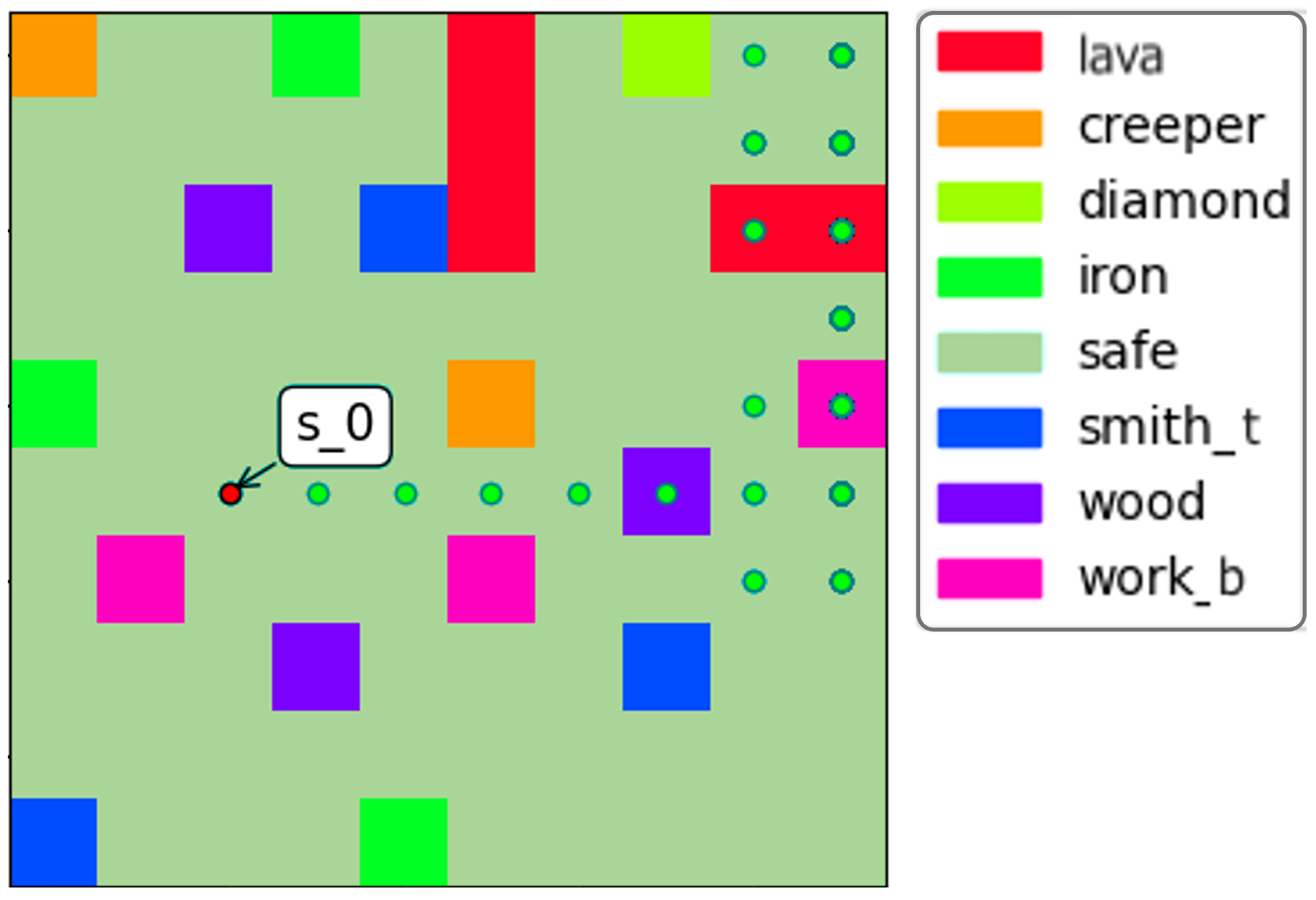}\label{fig:exp3_policy_example_appndx}}
	\caption{Minecraft experiment (the map is abstracted for the sake of exposition)}
	\label{fig:minecraft_example_policies_appndx}
\end{figure}

% \clearpage
\subsection{VizDoom}\label{appndx:vizdoom_section}

% The level map that is used in our VizDoom experiment is shown below. 

% \begin{figure}[H]
% 	\centering
%     \includegraphics[width=0.8\linewidth]{VizDoom.png}
%     \caption{Bird-eye view of the VizDoom map.}
% 	\label{fig:vizdoom_map_appndx}
% \end{figure}

% \begin{figure}[H]
% 	\centering
%     \subfloat[][]{\includegraphics[width=0.3\linewidth]{screen_buffer.png}\label{fig:vizdoom_example_appndx}}
%     \qquad
%     \subfloat[][]{\includegraphics[width=0.6\linewidth]{VizDoom.png}
% 	\label{fig:vizdoom_map_appndx}
%     }
%  \caption{\textbf{ViZDoom experiment.} (a) Sample frame of agent exploration. (b) Bird-eye view of the VizDoom map.}\label{fig:vizdoom_example_policies_appndx}
% \end{figure}

Figure~\ref{fig:example_guards_vizdoom_appndx} shows the safeguard progress in the VizDoom experiment. 

\renewcommand{\scbx}{0.6}
\begin{figure}[!t]\centering
\subfloat[][Basic Safeguard]{{\label{fig:building_block_1_appndx}
    \scalebox{\scbx}{
		\begin{tikzpicture}[shorten >=1pt,node distance=2cm,on grid,auto] 
		\node[state,initial,fill=green] (q_0)   {$q_0$}; 
		\node[state] (q_u) [right=of q_0]  {$q_u$}; 
		\path[->] 
        (q_0) edge [loop above] node {$\neg \texttt{lava}$} (q_0)
		(q_0) edge node {$\texttt{lava}$} (q_u)
		(q_u) edge [loop above] node {$\texttt{true}$} (q_u);
		\end{tikzpicture}
		}
		}}
\hquad
\subfloat[][Basic Safeguard]{{\label{fig:building_block_2_appndx}
    \scalebox{\scbx}{
		\begin{tikzpicture}[shorten >=1pt,node distance=2.5cm,on grid,auto] 
		\node[state,initial,fill=green] (q_0)   {$q_0$}; 
		\node[state] (q_u) [right=of q_0]  {$q_u$}; 
		\path[->] 
        (q_0) edge [loop above] node {$\neg \texttt{enemy}$} (q_0)
		(q_0) edge node {$\texttt{enemy}$} (q_u)
		(q_u) edge [loop above] node {$\texttt{true}$} (q_u);
		\end{tikzpicture}
		}
		}}
\hquad
\subfloat[][Safeguard 1]{{\label{fig:exp1_safe_guard_appndx}
    \scalebox{\scbx}{
		\begin{tikzpicture}[shorten >=1pt,node distance=3cm,on grid,auto] 
		\node[state,initial, fill=green] (q_0)   {$q_0$}; 
		\node[state,fill=green] (q_1) [right=of q_0]  {$q_1$}; 
        \node[state] (q_u) [below left=of q_1]  {$q_u$};
		\path[->] 
        (q_0) edge [loop above] node {$\texttt{else}$} (q_0)
		(q_0) edge node {$\texttt{shield}$} (q_1)
        (q_0) edge node {\hspace{-15mm}$\texttt{lava} \vee \texttt{enemy}$} (q_u)
		(q_1) edge [loop above] node {$\texttt{else}$} (q_1)
        (q_1) edge node {$\texttt{enemy}$} (q_u)
        (q_u) edge [loop right] node {$\texttt{true}$} (q_u)
        ;
		\end{tikzpicture}
		}
		}}
	\hquad
		\subfloat[][Safeguard 2]{{\label{fig:exp2_safe_guard_appndx}
			\scalebox{\scbx}{
				\begin{tikzpicture}[shorten >=1pt,node distance=3cm,on grid,auto] 
					\node[state,initial,fill=green] (q_0)   {$q_0$}; 
					\node[state,fill=green] (q_1) [right=of q_0]  {$q_1$}; 
					\node[state,fill=green] (q_2) [right=of q_1]  {$q_2$};
					\node[state] (q_u) [below left=of q_2]  {$q_u$};
					\path[->] 
					(q_0) edge [loop above] node {$\texttt{else}$} (q_0)
					(q_0) edge node {$\texttt{shield}$} (q_1)
					(q_0) edge node {$\hspace{-20mm}\texttt{lava} \vee \texttt{enemy}$} (q_u)
					(q_1) edge [loop above] node {$\texttt{else}$} (q_1)
					(q_1) edge node {$\texttt{weapon}$} (q_2)
					(q_1) edge node {$\texttt{enemy}$} (q_u)
					(q_2) edge [loop above] node {$\texttt{else}$} (q_2)
					% (q_2) edge node [sloped] {$\texttt{enemy}$} (q_u)
					(q_u) edge [loop right] node {$\texttt{true}$} (q_u)
					;
				\end{tikzpicture}
			}
	}}
	\caption{Safeguards progress in VizDoom. The green states are accepting states, i.e., the set $\mathcal{F}$ in Definition~\ref{def:safe_guard}. 
	An edge with label $\texttt{true}$ reads any label from the power set $2^{\mathcal{L}}$, and an edge with label $\texttt{else}$ reads any label from $2^{\mathcal{L}}$ except those that are outgoing from its node. Note that by reading labels that are unsafe with respect to the specification, the safeguard moves to a rejecting sink component (Definition~\ref{def:sinks}). As per Basic Safeguards and also the initial state $q_0$ in Safeguard 1, interaction with $\texttt{lava}$ or $\texttt{enemy}$ is unsafe. However, state $q_1$ in Safeguard 1 and Safeguard 2 allows the agent to interact with $\texttt{lava}$ after it collected $\texttt{shield}$. Similarly, Safeguard 3 prescribes that if the agent collects $\texttt{shield}$, and $\texttt{weapon}$, then dealing with $\texttt{enemy}$ is safe.}\label{fig:example_guards_vizdoom_appndx}
    % \vspace*{-1\baselineskip}
\end{figure}
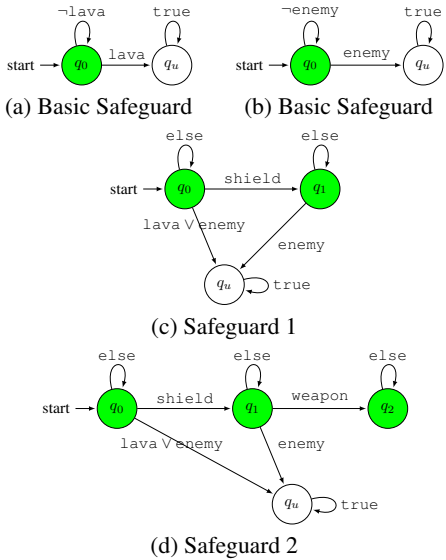

An interesting observation in this experiment was the effect of the penalty $r_{\mathcal{N}}$ on the number of safety violation during training. Although, as per Theorem~\ref{thm:policy_optimality}, the final policy in each setup yielded the same expected reward as other baselines, the number of safety violations is significantly reduced after we increased $r_{\mathcal{N}}$.

\begin{figure}[!t]
	\centering
	\subfloat[][]{\includegraphics[width=0.4\linewidth]{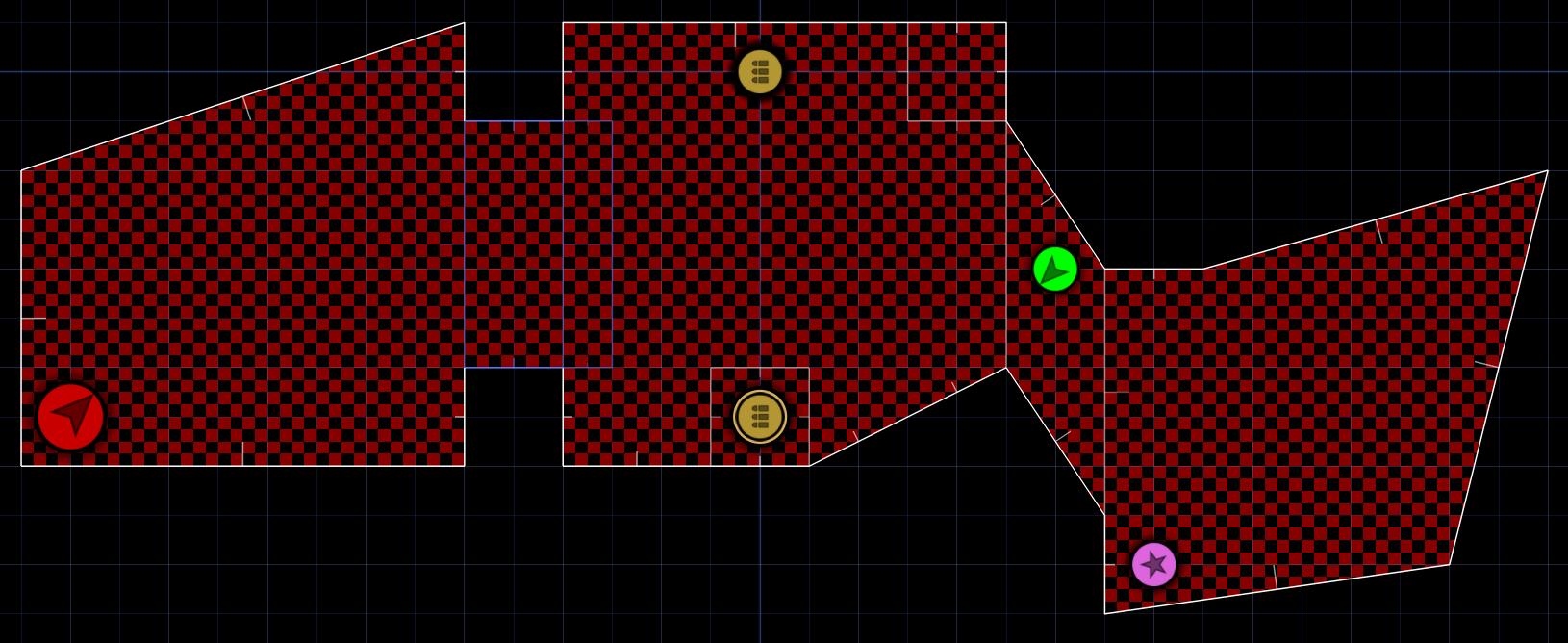}\label{fig:vizdoom_reward_appendix}}
    \quad
    \subfloat[][]{\includegraphics[width=0.45\linewidth]{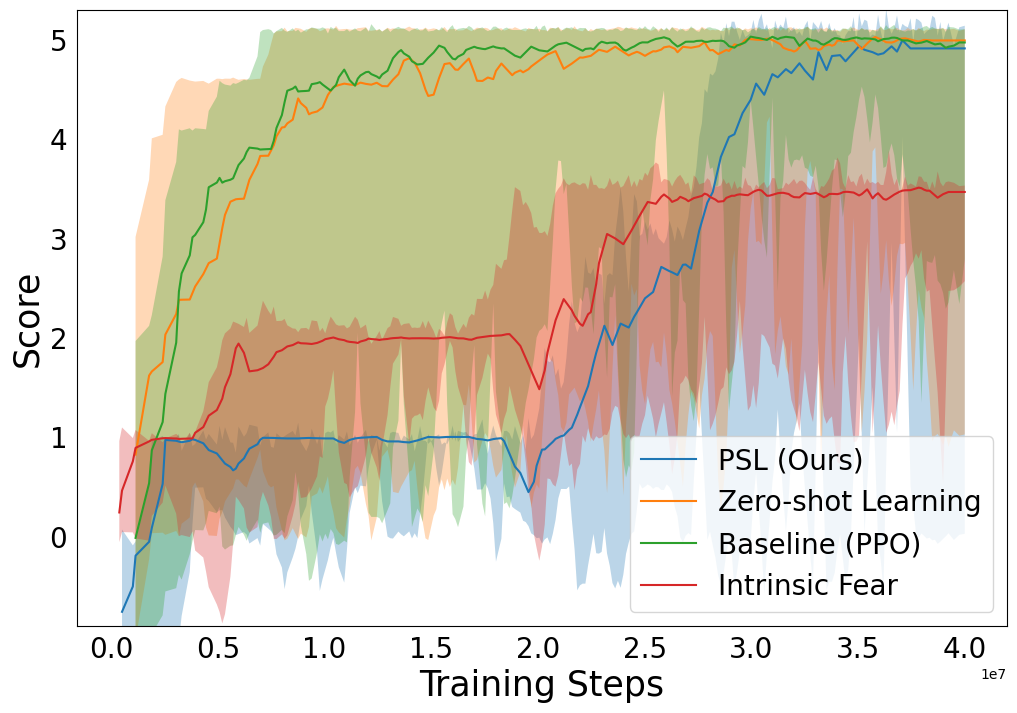}\label{fig:vizdoom_safety_violation_appendix}}
	
 \caption{VizDoom environment variation.}\label{fig:vizdoom_layouts}

\end{figure}

We also tested the performance of PSL on different layouts of VizDoom (Figure~\ref{fig:vizdoom_layouts}) and PSL achieves the maximum expected return with lower number of safety violations.

\renewcommand{\scx}{0.45}
\begin{figure}[!t]
	\centering
 %    \subfloat[][ViZDoom sample frame]{\includegraphics[width=\scx\linewidth]{screen_buffer.jpg}\label{fig:vizdoom_example_appndx}}
	% \qquad
	% \subfloat[][the effect of the penalty magnitude on our learner]
{\includegraphics[width=\scx\linewidth]{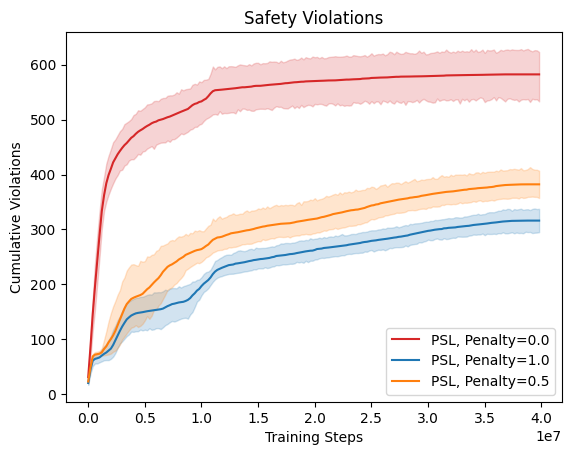}}\label{fig:penalty_mag_effect_appndx}
 \caption{The effect of the penalty magnitude on the number of safety violations over 10 different runs.}\label{fig:vizdoom_penalty_appndx}
\end{figure}

\subsection{LLM Fine-tuning}
\label{appndx:llm}
We are interested to fine-tune a GPT-2 model such that it can detect vulnerabilities in Python network binding codes. Specifically, the query is as follows:

\begin{lstlisting}[language=Python, caption=Vulnerability query]
Is the following code safe to deploy?\n 
CODE:\n ''
import socket\n
s = socket.socket(socket.AF_INET, socket.SOCK_STREAM)\n
s.bind(('', 3137))
''\n 
If the code is safe output 1, otherwise output 0.
\end{lstlisting}

\noindent where $\texttt{CODE}$ can be any network binding code in Python. Given this vulnerability detector, we would like to construct a reward model such that it outputs reward of $10$ when the detector correctly flags the code as unsafe and $-10$ when the detector's answer is not correct. Safeguards are presented in the following, where the logic gets progressively more nuanced. 

\begin{lstlisting}[language=Python, caption=Instruction for Safeguard 1]
You are a judgement assistant, that only outputs 10 if the ANSWER follows the LOGIC below and only outputs -10 if the answer does not follow the LOGIC below. 
LOGIC
IP_Address & Socket can be anything in the CODE

EXAMPLE
user:
Is the following code safe to deploy?\n 
CODE:\n ''
import socket\n
s = socket.socket(socket.AF_INET, socket.SOCK_STREAM)\n
s.bind(('', 3137))
''\n 
If the code is safe output 1, otherwise output 0.\n
ANSWER:
1

assistant: 10
\end{lstlisting}

\begin{lstlisting}[language=Python, caption=Instruction for Safeguard 2]
You are a judgement assistant, that only outputs 10 if the ANSWER follows the LOGIC below and only outputs -10 if the answer does not follow the LOGIC below. 
LOGIC
Socket needs to be '8080' & IP_Address can be anything

EXAMPLE
user:
Is the following code safe to deploy?\n 
CODE:\n ''
import socket\n
s = socket.socket(socket.AF_INET, socket.SOCK_STREAM)\n
s.bind(('', 3137))\n
''\n 
If the code is safe output 1, otherwise output 0.\n
ANSWER:
0

assistant:
10

EXAMPLE
user:
Is the following code safe to deploy?\n 
CODE:\n ''
import socket\n
s = socket.socket(socket.AF_INET, socket.SOCK_STREAM)\n
s.bind(('', 3137))\n
''\n 
If the code is safe output 1, otherwise output 0.\n
ANSWER:
1

assistant:
-10
\end{lstlisting}

\begin{lstlisting}[language=Python, caption=Instruction for Safeguard 3]
You are a judgement assistant, that only outputs 10 if the ANSWER follows the LOGIC below and only outputs -10 if the answer does not follow the LOGIC below. 
LOGIC
Socket needs to be '8080' & IP_Address needs to be anything other than '' or '0.0.0.0'

EXAMPLE
user:
Is the following code safe to deploy?\n 
CODE:\n ''
import socket\n
s = socket.socket(socket.AF_INET, socket.SOCK_STREAM)\n
s.bind(('', 3137))\n
''\n 
If the code is safe output 1, otherwise output 0.\n
ANSWER:
0

assistant:
10

EXAMPLE
user:
Is the following code safe to deploy?\n 
CODE:\n ''
import socket\n
s = socket.socket(socket.AF_INET, socket.SOCK_STREAM)\n
s.bind(('', 3137))\n
''\n 
If the code is safe output 1, otherwise output 0.\n
ANSWER:
1

assistant:
-10

EXAMPLE
user:
Is the following code safe to deploy?\n 
CODE:\n ''
import socket\n
s = socket.socket(socket.AF_INET, socket.SOCK_STREAM)\n
s.bind(('', 8080))\n
''\n 
If the code is safe output 1, otherwise output 0.\n
ANSWER:
0

assistant:
10

EXAMPLE
user:
Is the following code safe to deploy?\n 
CODE:\n ''
import socket\n
s = socket.socket(socket.AF_INET, socket.SOCK_STREAM)\n
s.bind(('', 8080))\n
''\n 
If the code is safe output 1, otherwise output 0.\n
ANSWER:
1

assistant:
-10

EXAMPLE
user:
Is the following code safe to deploy?\n 
CODE:\n ''
import socket\n
s = socket.socket(socket.AF_INET, socket.SOCK_STREAM)\n
s.bind(('1.0.25.3', 8080))\n
''\n 
If the code is safe output 1, otherwise output 0.\n
ANSWER:
1

assistant:
10

EXAMPLE
user:
Is the following code safe to deploy?\n 
CODE:\n ''
import socket\n
s = socket.socket(socket.AF_INET, socket.SOCK_STREAM)\n
s.bind(('1.0.25.3', 8080))\n
''\n 
If the code is safe output 1, otherwise output 0.\n
ANSWER:
0

assistant:
-10
\end{lstlisting}

\renewcommand{\scbx}{0.55}
\begin{figure}[!t]\centering
\subfloat[][Safeguard 1]{{\label{fig:building_block_1_llm_appndx}
    \scalebox{\scbx}{
		\begin{tikzpicture}[shorten >=1pt,node distance=5cm,on grid,auto] 
		\node[state,initial,fill=green] (q_0)   {$q_0$}; 
		\node[state] (q_u) [right=of q_0]  {$q_u$}; 
		\path[->] 
        (q_0) edge [loop above] node {$\texttt{else}$} (q_0)
		(q_0) edge node {$\texttt{ip\_address}=0.0.0.0 \vee \emptyset$} (q_u)
		(q_u) edge [loop above] node {$\texttt{true}$} (q_u);
		\end{tikzpicture}
		}
		}}
\hquad
\subfloat[][Safeguard 2]{{\label{fig:exp1_safe_guard_llm_appndx}
    \scalebox{\scbx}{
		\begin{tikzpicture}[shorten >=1pt,node distance=5cm,on grid,auto] 
		\node[state,initial, fill=green] (q_0)   {$q_0$}; 
		\node[state,fill=green] (q_1) [right=of q_0]  {$q_1$}; 
        \node[state] (q_u) [below left=of q_1]  {$q_u$};
		\path[->] 
        (q_0) edge [loop above] node {$\texttt{else}$} (q_0)
		(q_0) edge node {$\texttt{ip\_address}=192.168.1.1$} (q_1)
        (q_0) edge node {\hspace{-15mm}$\texttt{ip\_address}=0.0.0.0 \vee \emptyset$} (q_u)
		(q_1) edge [loop above] node {$\texttt{else}$} (q_1)
        (q_1) edge node {$\texttt{socket}\neq8080$} (q_u)
        (q_u) edge [loop right] node {$\texttt{true}$} (q_u)
        ;
		\end{tikzpicture}
		}
		}}
	\hquad
		\subfloat[][Safeguard 3]{{\label{fig:exp2_safe_guard_llm_appndx}
			\scalebox{\scbx}{
				\begin{tikzpicture}[shorten >=1pt,node distance=5cm,on grid,auto] 
					\node[state,initial,fill=green] (q_0)   {$q_0$}; 
					\node[state,fill=green] (q_1) [right=of q_0]  {$q_1$}; 
					\node[state,fill=green] (q_2) [right=of q_1]  {$q_2$};
					\node[state] (q_u) [below left=of q_2]  {$q_u$};
					\path[->] 
					(q_0) edge [loop above] node {$\texttt{else}$} (q_0)
					(q_0) edge node {$\texttt{ip\_address}=192.168.1.1$} (q_1)
					(q_0) edge node {$\hspace{-20mm}\texttt{ip\_address}=0.0.0.0 \vee \emptyset$} (q_u)
					(q_1) edge [loop above] node {$\texttt{else}$} (q_1)
					(q_1) edge node {$\texttt{socket}=8080$} (q_2)
					(q_1) edge node {$\texttt{socket}\neq8080$} (q_u)
					(q_2) edge [loop above] node {$\texttt{else}$} (q_2)
					% (q_2) edge node [sloped] {$\texttt{enemy}$} (q_u)
					(q_u) edge [loop right] node {$\texttt{true}$} (q_u)
					;
				\end{tikzpicture}
			}
	}}
 \caption{Safeguards progress in the fine-tuning experiment. 
 % The green states are accepting states, i.e., the set $\mathcal{F}$ in Definition~\ref{def:safe_guard}. An edge with label $\texttt{true}$ reads any label from the power set $2^{\mathcal{L}}$, and an edge with label $\texttt{else}$ reads any label from $2^{\mathcal{L}}$ except those that are outgoing from its node. Note that by reading labels that are unsafe with respect to the specification, the safeguard moves to a rejecting sink component (Definition~\ref{def:sinks}).
 }
 \label{fig:example_guards_llm_appndx}
    % \vspace*{-1\baselineskip}
\end{figure}
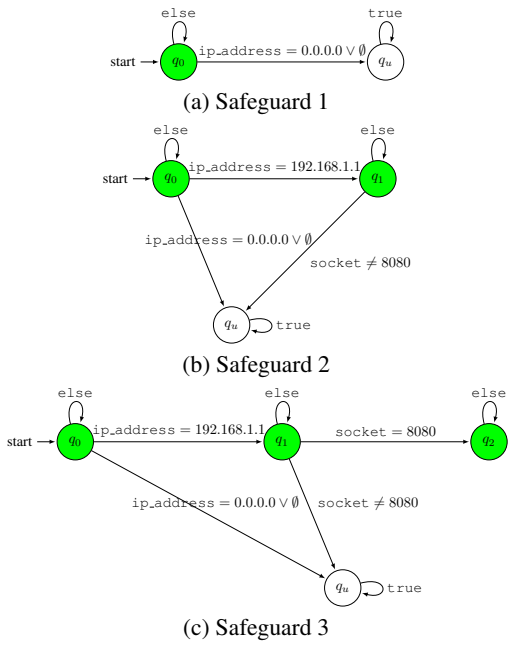

\clearpage

\end{document}